\pdfoutput=1

\documentclass[11pt]{article}

\usepackage[final]{acl}
\usepackage{times}
\usepackage{latexsym}
\usepackage[T1]{fontenc}
\usepackage[utf8]{inputenc}
\usepackage{microtype}
\usepackage{inconsolata}
\usepackage{graphicx}
\usepackage{amsmath}
\usepackage{amssymb}
\usepackage{mathtools}
\usepackage{amsthm}
\usepackage{algorithm}
\usepackage{algorithmic}
\usepackage{multirow}
\usepackage[capitalize,noabbrev]{cleveref}
\usepackage[dvipsnames]{xcolor}
\usepackage{enumitem}
\usepackage{booktabs}

\title{\vspace*{-0.5in}
{{\small \hfill EMNLP'25}\\
\vspace*{.25in}}\textit{\textbf{SimMark}}: A Robust Sentence-Level Similarity-Based \\ Watermarking Algorithm for Large Language Models}

\author{Amirhossein Dabiriaghdam \and Lele Wang \\
        Department of ECE, University of British Columbia, Vancouver, BC, Canada \\
        \texttt{\{amirhossein, lelewang\}@ece.ubc.ca}
        }

\begin{document}
\maketitle
\begin{abstract}
The widespread adoption of large language models (LLMs) necessitates reliable methods to detect LLM-generated text. We introduce \textbf{\textit{SimMark}}, a robust sentence-level watermarking algorithm that makes LLMs' outputs traceable without requiring access to model internals, making it compatible with both open and API-based LLMs. By leveraging the similarity of semantic sentence embeddings combined with rejection sampling to embed detectable statistical patterns imperceptible to humans, and employing a \textit{soft} counting mechanism, \textit{SimMark} achieves robustness against paraphrasing attacks. Experimental results demonstrate that \textit{SimMark} sets a new benchmark for robust watermarking of LLM-generated content, surpassing prior sentence-level watermarking techniques in robustness, sampling efficiency, and applicability across diverse domains, all while maintaining the text quality and fluency.\footnote{The source code of our algorithm is available \href{https://github.com/DabiriAghdam/SimMark}{here}.}
\end{abstract}

\section{Introduction}\label{intro}
The advent of deep generative models has made it increasingly important to determine whether a given text, image, or video was produced by artificial intelligence (AI), and recently, researchers across various domains have begun tackling this challenge \cite{aaronson2022, Fernandez_2023_ICCV,teymoorianfard2025vidstamptemporallyawarewatermarkownership}. In particular, LLMs such as GPT-4o \cite{hurst2024gpt}, can now generate human-like text at scale and low cost, enabling powerful applications across numerous industries and areas such as health care and law \cite{singhal2023large, wu2025medreason, torabi2025large, yao2024lawyer, taranukhin-etal-2024-empowering}.

This capability, however, introduces serious risks, including academic plagiarism, disinformation campaigns, and public opinion manipulation. For instance, the use of AI-generated content in news articles has raised concerns about transparency, accountability, and the spread of false information \cite{futurism2023cnet}. Moreover, reliably detecting LLM-generated content is crucial for enforcing copyright protections and ensuring accountability \cite{weidinger2021ethical}.

\begin{figure}[!t]
    \centering
    \includegraphics[page=1,width=\linewidth]{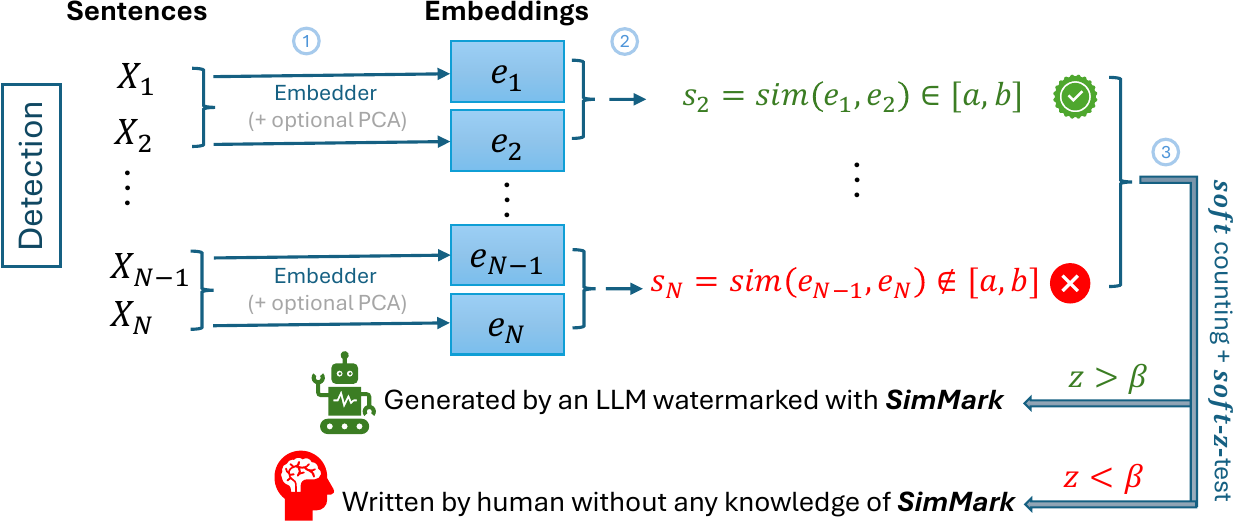}
    \caption{A high-level overview of \textit{SimMark} detection algorithm. The input text is divided into individual sentences $X_1$ to $X_N$, which are embedded using a semantic embedding model. 
    The similarity between consecutive sentence embeddings is computed. Sentences with similarities within a predefined interval $[a, b]$ are considered \textcolor{green!50!black}{\textbf{valid}}, while those outside are \textcolor{red}{\textbf{invalid}}. 
    A statistical test is performed using the count of \textcolor{green!50!black}{\textbf{valid}} sentences to determine whether the text is watermarked.}
    \label{fig:Detection}
\end{figure} 

Detecting LLM-generated text poses a unique challenge. These models are explicitly trained to emulate human writing styles, often rendering their outputs indistinguishable from human-authored text. As demonstrated by \citet{kumarage-etal-2023-reliable} and \citet{reliable}, reliably differentiating between human-written and machine-generated text remains an open problem.

One promising approach is the use of imperceptible statistical signatures, or \textit{watermarks}, embedded within a text. Watermarking imperceptibly alters text such that it remains natural to human readers but enables subsequent detection of its origin \cite{atallah2001natural}. Effective watermarking must balance the preservation of the text quality with robustness against adversarial \textit{paraphrasing}, where attackers modify the text to evade detection \cite{krishna2024paraphrasing}. Additionally, watermarks must be resistant to \textit{spoofing} attacks, wherein adversaries craft non-machine-generated text (often malicious) to falsely trigger detectors \cite{reliable}.

In this paper, we introduce \textbf{\textit{SimMark}}, a robust sentence-level watermarking algorithm for LLMs based on sentence embedding similarity. \textit{SimMark} treats LLMs as black boxes that can be prompted to generate sentences given a context. This approach makes \textit{SimMark} compatible with a wide range of models, including open-weight LLMs and closed-source proprietary models accessible only via APIs, as it does not require fine-tuning or access to the models' internal logits. Access to logits is often restricted by API providers due to their potential use in distilling LLMs and leaking proprietary information \cite{finlayson2024logits}. 

\textit{SimMark} leverages embeddings from semantic text embedding models to capture semantic relationships between sentences and embeds detectable statistical patterns on sentence similarity through rejection sampling. Specifically, rejection sampling involves querying the LLM multiple times until the similarity between the embeddings of consecutive sentences falls within a predefined interval. During detection, these patterns are analyzed using a statistical test to differentiate between human-written and LLM-generated text, as illustrated in Figure~\ref{fig:Detection}.

In summary, our contributions are as follows:
\begin{itemize}[noitemsep, topsep=0pt]
    \item We introduce a novel sentence-level watermarking method that achieves state-of-the-art detection performance while maintaining low false positive rates for human-written text.
    \item Our approach demonstrates robustness against paraphrasing attacks through semantic-level watermarking and a soft counting mechanism for statistical testing.
    \item Compared to existing methods, \textit{SimMark} provides a more practical solution that operates without access to LLM logits, offering high-quality watermark injection and detection.
\end{itemize}

The remainder of this paper is organized as follows: Section~\ref{background} reviews the background and related work on LLM watermarking techniques. Section~\ref{approach} outlines our methodology. Section~\ref{experiments} describes our experimental setup and presents comparative results, while Section~\ref{conclusion} concludes the paper.

\section{Background} \label{background}
In this section, we provide an overview of the foundational concepts related to text generation with LLMs and discuss related works on watermarking techniques for LLMs.

\subsection{Autoregressive Decoding of LLMs}  
An LLM operates over a vocabulary \( V \), a set of words or subwords termed as \textit{tokens}. Let \( f: V \rightarrow V \) be an LLM that takes a sequence of tokens \( T_i = \{t_1, t_2, \ldots, t_i\} \) as input and generates the next token \( t_{i+1} \) as its output. To generate \( t_{i+1} \), the LLM samples it from the conditional probability distribution \( P(t_{i+1} | T_i) \) over the vocabulary \( V \). After generating \( t_{i+1} \), the updated sequence \( T_{i+1} = T_i \cup \{t_{i+1}\} \) is fed back into the model, and the process is repeated iteratively to generate the subsequent tokens. This process of generating one token at a time, given the previously generated tokens, is known as \textit{autoregressive decoding}.  

\subsection{Token-Level Watermarking}  
Token-level watermarking methods embed a statistical signal in the text by manipulating the token sampling process \cite{aaronson2022, kgw, fu2024watermarking}. These methods typically alter the probability distribution over \( V \), subtly biasing the selection of certain tokens to form detectable patterns.

KGW introduced by \citet{kgw}, groups \( V \) into \textit{green} and \textit{red} subsets \textit{pseudo-randomly} seeded on the previous token before generating each new token. A predefined constant \(\delta > 0\) is added to the logits of each token in the green list, increasing their likelihood of being selected during the sampling step. At detection, a \( z \)-test is applied to the number of tokens from the green list in the text to determine whether the text contains a watermark. This test compares the observed proportion of green tokens to the expected proportion under the null hypothesis of no watermark, providing a statistical measure to detect even subtle biases introduced by the watermark.

Detection of such watermarks involves analyzing tokens for statistical signatures that deviate from typical human text. However, token-level watermarks can still be vulnerable to paraphrasing, as rephrasing may disrupt the green and red token lists without altering the overall semantic \cite{krishna2024paraphrasing}. Moreover, since these methods modify the logits, they directly impact the conditional probability distribution over \( V \), potentially degrading the quality of the generated text \cite{fu2024watermarking}.

Due to space constraints, additional related work—including token-level methods such as UNIGRAM-WATERMARK (UW) \cite{unigram} and the Semantic Invariant Robust (SIR) watermark \cite{sir}, as well as post-hoc watermarking techniques—is deferred to Appendix~\ref{appendix:additional-related}.

\subsection{Sentence-Level Watermarking}  
One approach to mitigate the previously mentioned problems is to inject the watermark signal at the sentence level, making it less vulnerable to adversarial modifications \cite{topkara2006words}. Consider a similar notation for sentence generation using an autoregressive LLM that takes a sequence of sentences \( M_i = \{X_1, X_2, \ldots, X_i\} \) and generates the next sentence \( X_{i+1} \). The updated sequence of sentences \( M_{i+1} = M_i \cup \{X_{i+1}\} \) is then used to generate subsequent sentences iteratively.

SemStamp by \citet{hou-etal-2024-semstamp} employs Locality-Sensitive Hashing (LSH)~\cite{lsh} to pseudo-randomly partition the semantic space of an embedding model into a set of \textit{valid} and \textit{blocked} regions, analogous to the green and red subsets in KGW. During rejection sampling, if the embedding of a newly generated sentence lies within the valid regions (determined based on the LSH signature of the previous sentence), the sentence is accepted. Otherwise, a new sentence is generated until a valid sentence is produced or the retry limit is reached. Similar to KGW, a \( z \)-test is applied to the number of valid sentences to determine whether the text contains a watermark.  

To improve robustness against paraphrasing, SemStamp used a contrastive learning approach \cite{1640964}, fine-tuning an embedding model such that the embeddings of paraphrased sentences remain as close as possible to the original sentences. This was achieved by minimizing the distance between paraphrased and original embeddings while ensuring unrelated sentences remained distinct. They also introduce a margin constraint in the rejection sampling process to reject sentences whose embeddings lie near the region boundaries.

$k$-SemStamp~\cite{{hou-etal-2024-k}} builds upon SemStamp and aims to enhance robustness by partitioning the semantic space using \( k \)-means clustering~\cite{kmeans} instead of random partitioning. They claim that in this way, sentences with similar semantics are more likely to fall within the same partition, unlike random partitioning, which may place semantically similar sentences into different partitions, reducing robustness; however, $k$-SemStamp assumes that the LLM generates text within a specific domain to apply \( k \)-means clustering effectively \cite{{hou-etal-2024-k}}, limiting its applicability in real-world, open-domain scenarios. The generation and detection procedures of $k$-SemStamp remain similar to the original SemStamp. In contrast to token-level algorithms, these sentence-level methods do not alter the internals of the LLM, therefore, it is expected that their output to be of higher quality \cite{hou-etal-2024-semstamp, hou-etal-2024-k}.

Our work, similar to SemStamp and $k$-SemStamp, is a sentence-level algorithm; however, it injects its watermark signature into the semantic similarity of consecutive sentences. It achieves great generalizability across domains by leveraging any off-the-shelf, general-purpose embedding model without fine-tuning. At the same time, it outperforms these state-of-the-art (SOTA) sentence-level watermarking methods in robustness against paraphrasing while preserving text quality.

\section{\textit{SimMark}: A Similarity-Based Watermarking Algorithm} 
\label{approach}
In this section, we present our proposed framework for watermarking LLMs, detailing both the process of generating watermarked text and its subsequent detection.

\begin{figure*}[t]
    \centering
    \includegraphics[page=2,width=\linewidth]{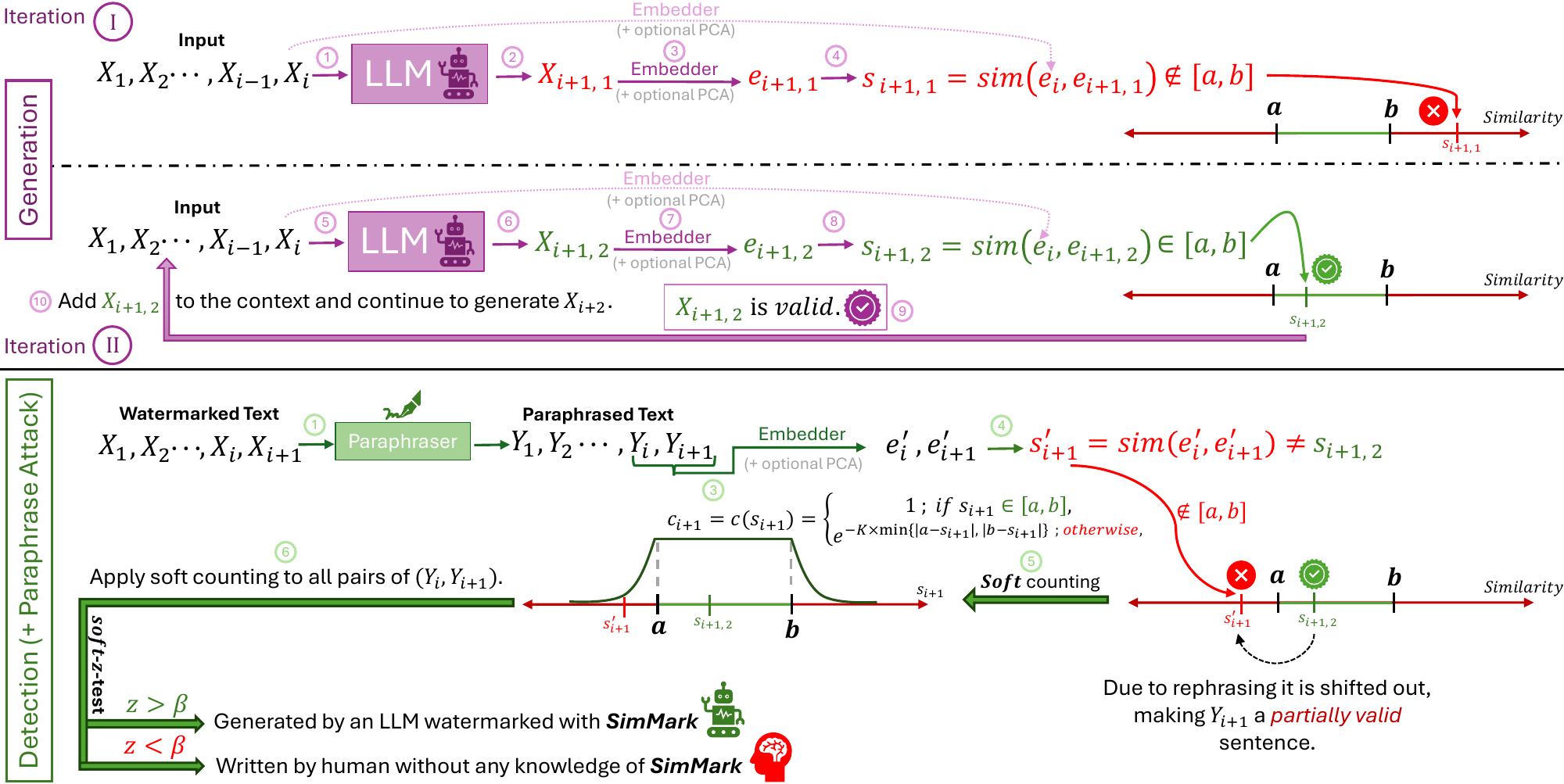} 
    \caption{Overview of \textit{\textbf{SimMark}}. \textbf{Top:} \textit{\textbf{Generation}}. For each newly generated sentence ($X_{i+1}$), its embedding ($e_{i+1}$) is computed using a semantic text embedding model, optionally applying PCA for dimensionality reduction. The cosine similarity (or Euclidean distance) between $e_{i+1}$ and the embedding of the previous sentence ($e_i$), denoted as $s_{i+1}$, is calculated. If $s_{i+1}$ lies within the predefined interval $[a, b]$, the sentence is marked \textcolor{green!50!black}{valid} and accepted. Otherwise, rejection sampling generates a new candidate sentence until validity is achieved or the iteration limit is reached. Once a sentence is accepted, the process repeats for subsequent sentences. \textbf{Bottom:} \textit{\textbf{Detection (+ Paraphrase attack)}}. Paraphrased versions of watermarked sentences are generated ($Y_{i}$), and their embeddings ($e'_{i}$) are computed. The similarity between consecutive sentences in the paraphrased text is evaluated. If paraphrasing causes the similarity ($s'_{i+1}$) to fall outside $[a, b]$, it is mismarked as \textcolor{red}{invalid}. A \textit{soft counting} mechanism (via function $c(s_{i+1})$ instead of a regular counting with a step function in $[a,b]$) quantifies partial validity based on proximity to the interval bounds, enabling detection of watermarked text via a \textit{\textbf{soft}}-$z$-test even under paraphrase attacks. It should be noted that soft counting is always applied, as we cannot assume prior knowledge of paraphrasing.}
    \label{fig:Overview}
\end{figure*}

\subsection{Watermarked Text Generation}
Similar to SemStamp and $k$-SemStamp, \textit{SimMark} utilizes the embedding representations of the sentences. To compute the embeddings, in contrast with \citet{hou-etal-2024-semstamp, hou-etal-2024-k} that fine-tuned their embedder model (which could make it biased toward a specific paraphrasing model or domain), we employ Instructor-Large~\cite{su-etal-2023-one}, a general-purpose embedding model, without any fine-tuning. The flexibility of our method in using any pretrained embedding model enables our approach to be more easily adaptable to different domains. 

\begin{algorithm}[t]
    \caption{\textit{{SimMark}} Generation Pseudo-Code}
    \label{alg:generation}
    \begin{algorithmic}[1]
       \REQUIRE LLM, embedding model, interval $[a, b]$, maximum iterations $N_\mathrm{max}$
        \FOR{each generated sentence $X_i$}
            \STATE Compute embedding $e_i$ for $X_i$ using the embedding model.
            \STATE $n \gets 0$ 
            \STATE \textbf{do}
            \STATE \hspace{1em} Generate sentence $X_{i+1}$ using the LLM.
            \STATE \hspace{1em} Compute embedding $e_{i+1}$ for $X_{i+1}$ using the embedding model.
            \STATE \hspace{1em} \textit{Optional:} Reduce the dimension of $e_i$ and $e_{i+1}$ using the PCA model.
            \STATE \hspace{1em} Compute similarity $s_{i+1}$:
             {\small \[
            s_{i+1} \gets 
            \begin{cases}
                \tfrac{e_i\cdot e_{i+1}}{\Vert e_{i}\Vert_2 \cdot \Vert e_{i+1}\Vert_2} &  \text{ if cosine similarity}, \\
                \Vert e_i - e_{i+1}\Vert_2 & \text{ if Euclidean distance,}
            \end{cases}
            \]}
            \STATE \hspace{1em} $n \gets n + 1$
            \STATE \textbf{while} $s_{i+1} \textcolor{red}{\notin [a, b]}$ \textbf{and} $n < N_\mathrm{max}$
            \STATE Accept $X_{i+1}$ as valid and continue generating the next sentence. \COMMENT{\textit{Here we either reached the $N_\mathrm{max}$ or $s_{i+1} \textcolor{green!50!black}{\in [a, b]}$}}
        \ENDFOR
    \end{algorithmic}
\end{algorithm}

First, we compute the embedding for each sentence\footnote{In our experiments, we passed both the sentence and \textit{``Represent the sentence for cosine similarity:''} or \textit{``Represent the sentence for Euclidean distance:''} as the instruction to the Instructor-Large model.}. Then, we calculate the cosine similarity (or Euclidean distance) between the embedding of sentence $i+1$ and the embedding of sentence $i$. If the computed value lies within a predefined interval, sentence $i+1$ is considered \textbf{\textit{\textcolor{green!50!black}{valid}}} (analogous to the green subset in KGW). Otherwise, we prompt the LLM to generate a new sentence and repeat this procedure until a valid sentence is found or the maximum number of iterations is reached (in this case, we accept the last generated sentence), as shown in Algorithm~\ref{alg:generation}. Optionally, we may apply Principal Component Analysis (PCA) \cite{jolliffe2002principal} method to the embeddings to reduce their dimensionality before calculating the similarity\footnote{In our experiments, this is the instruction in this case: \textit{``Represent the sentence for PCA:''}}. The reason for applying PCA is provided in Subsection \ref{sub:detection}. An overview of \textit{SimMark} generation algorithm is depicted in the top part of Figure~\ref{fig:Overview}.

The predefined interval is a hyperparameter chosen a priori based on the distribution of similarities between consecutive sentences' embeddings generated by an unwatermarked LLM and human-written text. The choice of this hyperparameter is critical for the performance of \textit{SimMark}.

First, if the interval's width is too small or its position is far from the mean of the similarity distribution, generating sentences within this interval can be challenging or even infeasible for the LLM. Conversely, if the interval's width is large and centered around the mean of the similarity distribution, generating sentences becomes easier, but the false positive (FP) rate (i.e., human-written text misclassified as machine-generated) increases. 

Furthermore, the choice of interval affects the robustness of \textit{SimMark} against paraphrasing attacks. When an attacker paraphrases sentences, the similarities may change and fall outside the interval. Consequently, a larger interval provides greater robustness, as the watermark is less likely to be disrupted by paraphrasing. Therefore, the selection of the interval involves balancing several factors: it must not be too narrow to impede sentence generation while maintaining a low FP rate and adequate robustness against paraphrasing.

Finding a ``sweet spot,'' for the interval depends on the distribution of similarities between consecutive sentences, which can vary across models. However, identifying such sweet spots is feasible when we analyze the similarity distributions of both human-authored and LLM-generated text (see Appendix \ref{appendix:optimal_interval} for an example of finding a sweet spot).

\subsection{Watermarked Text Detection} \label{sub:detection}

The detection of \textit{SimMark} follows a similar methodology to KGW by employing a $z$-test for hypothesis testing. However, akin to \citet{hou-etal-2024-semstamp}, the detection operates at the \textit{sentence level} rather than the token level. To perform detection, we first divide the input text into sentences and use the same semantic embedding model to compute the embeddings for each sentence. If PCA was applied during the watermarking process, it must also be applied during detection to ensure consistency. 

\begin{algorithm}[t]
    \caption{\textit{{SimMark}} Detection Pseudo-Code}
    \label{alg:detection}
    \begin{algorithmic}[1]
       \REQUIRE input text, embedding model, interval $[a, b]$, decay factor $K$, threshold $\beta$
        \STATE Split the input text into sentences excluding the first sentence (i.e, the prompt): $M=\{X_2, \dots, X_{N_\mathrm{total}}\}$, and set $N \gets |M|$.
        \FOR{each sentence pair $(X_i, X_{i+1})$}
            \STATE Compute embeddings $e_i$ for $X_i$ and $e_{i+1}$ for $X_{i+1}$ using the embedding model.
            \STATE \textit{Optional}: Reduce the dimensionality of $e_i$ and $e_{i+1}$ using the PCA model.
            \STATE Compute $s_{i+1}\gets sim(e_i, e_{i+1})$.
            \STATE Compute $c_{i+1}$ according to Eq. \eqref{eq:count_formula}.
        \ENDFOR
        \STATE Soft count of valid sentences: $N_\mathrm{valid\_soft} \gets \sum_{i=2}^{N+1} c_i$. 
        \STATE Estimate $p_0$ as the area under the human-written text embeddings similarity distribution curve within $[a, b]$.
        \STATE Compute $z_\mathrm{soft}$ using Eq. \eqref{eq:z_formula2}.
        \IF{\textcolor{green!50!black}{$z_\mathrm{soft} > \beta$}} 
            \STATE Reject $H_0$, i.e., text is likely generated by an LLM watermarked by \textit{SimMark}. 
            ~\raisebox{-0.3\height}{\includegraphics[width=0.033\textwidth]{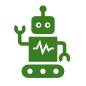}}
            
        \ELSE  
            \STATE Accept $H_0$, i.e., text is likely human-written.
        \ENDIF
    \end{algorithmic}
\end{algorithm}

Next, we compute the similarity ($s_{i+1}=sim(e_i,e_{i+1})$) between consecutive sentences ($X_{i}$ and $X_{i+1}$) and count the number of \textcolor{green!50!black}{valid} sentences ($N_\mathrm{valid\_soft}$). Sentence $X_{i+1}$ is deemed valid if $s_{i+1}$ lies within the predefined interval $[a, b]$. However, paraphrasing may alter embeddings significantly, causing the similarity to deviate from the desired interval. To mitigate this, we adopt a \textbf{soft counting} approach, where a sentence is considered \textit{partially valid} if its similarity is near the interval boundaries. Specifically, the soft count of $X_{i+1}$, denoted as $c_{i+1}$, is defined as follows:

{\small
\begin{equation}
\scalebox{0.9}{$
\begin{aligned}
c_{i+1} = c(s_{i+1}) = 
\begin{cases} 
1 & \text{if } s_{i+1} \; \textcolor{green!50!black}{\in [a, b]}, \\
e^{-K \min\{|a - s_{i+1}|, |b - s_{i+1}|\}} & \textcolor{red}{\text{otherwise.}}
\end{cases}
\end{aligned}
$}
\label{eq:count_formula}
\end{equation}
}
Here, \( K > 0 \) is a decay factor controlling the smoothness of the soft counting. A higher \( K \) makes the function behave closer to a step function, while a lower \( K \) allows for smoother transitions, tolerating minor deviations outside the interval \([a, b]\). The total number of valid sentences is then computed as $N_\mathrm{valid\_soft} = \sum_i c_i$. Refer to Appendix \ref{appendix:K} for an ablation study on how this approach improves robustness against paraphrasing, with only a minimal impact on performance in non-paraphrased scenarios, by allowing for some degree of error. 

During our initial experiments, we observed that, contrary to cosine similarity, Euclidean distance is very sensitive to paraphrasing, and even a subtle change in the sentences would result in a huge difference in the distances of embeddings. We hypothesize that Euclidean distance is sensitive to noise in high-dimensional spaces such as the semantic space of an embedder. To mitigate this, we propose using PCA: We fit a PCA model on a dataset of human-written texts to find the principal components of the sentence embeddings, i.e., the components that contribute the most to the semantical representation of the sentences. Then, we apply PCA to reduce embeddings' dimension. More details on the dimensionality reduction are provided in Section \ref{experiments} and Appendix \ref{appendix:ablation_pca}.

This approach can make reverse-engineering more difficult, as it would require knowledge of not only the embedder model, but also the PCA setting (e.g., number of components, access to the dataset used for fitting it, etc.). Without all these details, reproducing the similarity distribution becomes less straightforward.

The null hypothesis $H_0$ is defined as follows:
\begin{quote}
\textit{$H_0$: The sentences are written by humans, i.e., the text sequence is generated without knowledge of the valid interval in the similarity of sentence embeddings.}
\end{quote}
We calculate the $z$-statistic for the one-proportion $z$-test using the sample proportion $p = \frac{N_\mathrm{valid\_soft}}{N}$, where $N$ is the total number of samples (sentences). Since $N_\mathrm{valid\_soft}$ is a soft count of valid sentences, we refer to it as \textit{\textbf{soft}}-$z$-score, which is given by $z_\mathrm{soft}  = \frac{p - p_0}{\sqrt{\frac{p_0(1 - p_0)}{N}}}$ or alternatively: \begin{equation}
z_\mathrm{soft} = \tfrac{N_\mathrm{valid\_soft} - p_0 N}{\sqrt{p_0(1 - p_0)N}}.
\label{eq:z_formula2}
\end{equation}
Here, the population proportion $p_0$ represents the ratio of valid sentences to all sentences in human-written text (i.e., a text with no watermark), which is estimated as the area under the similarity distribution curve of consecutive human-written sentences within the interval $[a, b]$ (like the one in Figure~\ref{fig:dist_of_distances} in Appendix \ref{appendix:optimal_interval}). The value of $z_\mathrm{soft}$ can be interpreted as a normalized deviation of the number of valid sentences $N_\mathrm{valid\_soft}$ from its expectation $p_0 N$.

As highlighted in Algorithm~\ref{alg:detection}, the null hypothesis $H_0$ is rejected if \textcolor{green!50!black}{$z_{soft} > \beta$}, where $\beta$ is a threshold determined empirically by running the detection algorithm on human-written text. The threshold $\beta$ is selected to maintain a desired FP rate (i.e., minimizing the misclassification of human-written text as LLM-generated). Details on the computation of $\beta$ are provided in Appendix \ref{appendix:estimate-AUC}.

\section{Experiments \& Results} \label{experiments}
\renewcommand*{\thefootnote}{\fnsymbol{footnote}}
\begin{table*}[t]
\centering
\small
\resizebox{\textwidth}{!}{
\begin{tabular}{@{}clc|ccccccc@{}}
\toprule
Dataset & Algorithm & No Paraphrase & Pegasus & Pegasus-Bigram & Parrot & Parrot-Bigram & GPT3.5 & GPT3.5-bigram & Avg. Paraphrased \\ 
\midrule
\multirow{7}{*}{\rotatebox{90}{RealNews}} 
 & UW (\citeauthor{unigram}) & \textbf{99.9} / \underline{99.1} / \textbf{99.9} & 98.5 / 85.6 / 95.3 & 97.9 / 73.5 / 91.7 & \underline{97.9} / 70.9 / \underline{91.9} & 97.4 / 62.8 / \underline{89.4} & \textbf{97.4} / \underline{59.1} / \textbf{87.9} & \textbf{93.7} / 37.0 / \underline{70.8}  & \underline{97.1} / 64.8 / \underline{87.8} \\
& KGW (\citeauthor{kgw})        & 99.6 / 98.4 / 98.9 & 95.9 / 82.1 / 91.0 & 92.1 / 42.7 / 72.9 & 88.5 / 31.5 / 55.4 & 83.0 / 15.0 / 39.9 & 82.8 / 17.4 / 46.7 & 75.1 / 5.9 / 26.3 & 86.2 / 32.4 / 55.4 \\
& SIR (\citeauthor{sir})       & \textbf{99.9} / \textbf{99.4} / \textbf{99.9} & 94.4 / 79.2 / 85.4 & 94.1 / 72.6 / 82.6 & 93.2 / 62.8 / 75.9 & 95.2 / 66.4 / 80.2 & 80.2 / 24.7 / 42.7 & 77.7 / 20.9 / 36.4 & 89.1 / 54.4 / 67.2 \\
& SemStamp (\citeauthor{hou-etal-2024-semstamp}) & 99.2 / 93.9 / 97.1 & 97.8 / 83.7 / 92.0 & 96.5 / 76.7 / 86.8 & 93.3 / 56.2 / 75.5 & 93.1 / 54.4 / 74.0 & 83.3 / 33.9 / 52.9 & 82.2 / 31.3 / 48.7 & 91.0 / 56.0 / 71.6 \\
& $k$-SemStamp (\citeauthor{{hou-etal-2024-semstamp}}) & 99.6 / 98.1 / 98.7 & \textbf{99.5} / \textbf{92.7} / \underline{96.5} & \underline{99.0} / \underline{88.4} / \underline{94.3} & 97.8 / \underline{78.7} / 89.4 & \underline{97.5} / \underline{78.3} / 87.3 & 90.8 / 55.5 / 71.8 & 88.9 / \textbf{50.2} / 66.1  & 95.6 / \underline{74.0} / 84.2  \\ 
& \textbf{Cosine-\textit{SimMark} (ours)} & 99.6 / 96.8 / 98.8 & \underline{99.2} / \underline{90.3} / \textbf{98.2} & \textbf{99.1} / \textbf{90.3} /\textbf{ 97.9} & \textbf{98.7} / \textbf{88.1} / \textbf{97.2} & \textbf{98.8} / \textbf{87.3} / \textbf{97.6} & \underline{95.7} / \textbf{59.7} / \underline{86.7} & \underline{92.0} / \underline{38.8} / \textbf{73.7}  & \textbf{97.2} / \textbf{75.8} / \textbf{91.9} \\ 
& \textbf{Euclidean-\textit{SimMark}\footnotemark[7] (ours)} & \underline{99.8} / 98.5 / \underline{99.3} & 97.2 / 72.3 / 89.1 & 96.9 / 70.0 / 87.4 & 95.7 / 60.2 / 82.5 & 95.7 / 59.1 / 81.5 & 94.1 / 51.6 / 76.2 & 88.2 / 29.7 / 53.5  & 94.6 / 57.2 / 78.4 \\ 
\midrule
\multirow{7}{*}{\rotatebox{90}{BookSum}} 
 & UW  & \textbf{100} / \textbf{100} / \textbf{100} & \textbf{99.5} / 89.8 / \textbf{98.5} & 98.6 / 71.2 / 93.0 & \underline{98.9} / 79.4 / \underline{94.8} & 98.6 / 72.1 / 92.9 & 93.2 / 24.6 / 57.9 & 86.0 / 9.2 / 30.5  & 95.8 / 57.7 / 77.9 \\
& KGW   & 99.6 / 99.0 / 99.2 & 97.3 / 89.7 / 95.3 & 96.5 / 56.6 / 85.3 & 94.6 / 42.0 / 75.8 & 93.1 / 37.4 / 71.2 & 87.6 / 17.2 / 52.1 & 77.1 / 4.4 / 27.1  & 91.0 / 41.2 / 67.8 \\
& SIR         & \textbf{100} / \underline{99.8} / \textbf{100}   & 93.1 / 79.3 / 85.9 & 93.7 / 69.9 / 81.5 & 96.5 / 72.9 / 85.1 & 97.2 / 76.5 / 88.0 & 80.9 / 39.9 / 23.6 & 75.8 / 19.9 / 35.4  & 89.5 / 59.7 / 66.6 \\
& SemStamp    & 99.6 / 98.3 / 98.8 & 99.0 / \textbf{94.3} / 97.0 & 98.6 / 90.6 / 95.5 & 98.3 / 83.0 / 91.5 & 98.4 / 85.7 / 92.5 & 89.6 / 45.6 / 62.4 & 86.2 / 37.4 / 53.8  & 95.0 / 72.8 / 82.1 \\
& $k$-SemStamp & \underline{99.9} / 99.1 / 99.4 & \underline{99.3} / \underline{94.1} / \underline{97.3} & \underline{99.1} / \underline{92.5} / \underline{96.9} & 98.4 / \underline{86.3} / 93.9 & \underline{98.8} / \textbf{88.9} / \underline{94.9} & 95.6 / \underline{65.7} / 83.0 & \underline{95.7} / \underline{64.5} / \underline{81.4}  & 97.8 / \underline{81.5} / 91.2 \\ 
& \textbf{Cosine-\textit{SimMark} (ours) }& 99.8 / 98.8 / \underline{99.5} & \textbf{99.5} / 93.3 / \textbf{98.5} & \textbf{99.6} / \textbf{94.1} / \textbf{98.5} & \textbf{99.3} / \textbf{88.5} / \textbf{98.0} & \textbf{99.3} / \underline{87.0} / \textbf{98.2} & \underline{97.1} / 62.5 / \underline{86.9} & 94.5 / 41.6 / 74.2  & \underline{98.2} / 77.8 / \underline{92.4} \\ 
& \textbf{Euclidean-\textit{SimMark} (ours)} & \textbf{100} / \textbf{100} / \textbf{100} & 98.8 / 82.6 / 94.9 & 98.6 / 80.4 / 93.4 & 97.9 / 75.3 / 91.1 &  97.9 / 73.3 / 91.6 & \textbf{99.7} / \textbf{94.4} / \textbf{98.8} & \textbf{99.5} / \textbf{91.9} / \textbf{97.6}  & \textbf{98.7} / \textbf{83.0} / \textbf{94.6} \\ 
\midrule
\multirow{6}{*}{\rotatebox{90}{\small Reddit-TIFU}} 
 & UW  & \textbf{99.9} / \textbf{99.5} / \textbf{99.8} & 97.3 / 73.4 / 91.1 & 94.1 / 48.3 / 77.2 & 90.6 / 37.1 / 64.0 & 89.2 / 33.7 / 60.4 & 86.3 / 26.9 / 52.9 & 74.3 / 13.2 / 30.0  & 88.6 / 38.8 / 62.6 \\
& KGW & 99.3 / 97.5 / 98.1 & 94.1 / 87.2 / 87.2 & 91.7 / 67.2 / 67.6 & 79.5 / 22.8 / 43.3 & 82.8 / 27.6 / 49.7 & 84.1 / 27.3 / 50.9 & 79.8 / 19.3 / 41.3  & 85.3 / 41.9 / 56.7  \\
& SIR & 99.6 / 97.2 / \underline{99.7} & 90.0 / 48.7 / 77.4 & 90.9 / 33.1 / 71.1 & 87.1 / 15.0 / 50.9 & 86.9 / 12.8 / 49.8 & 91.1 / 15.0 / 61.4 & 84.3 / 5.5 / 39.1 & 88.4 / 21.7 / 58.3 \\
& SemStamp & 99.7 / 97.7 / 98.2 & 98.4 / 92.8 / 95.4 & 98.0 / 89.0 / 92.9 & 90.2 / 56.2 / 70.5 & 93.9 / 71.8 / 82.3 & 87.7 / 47.5 / 58.2 & 87.4 / 43.8 / 55.9  & 92.6 / 66.9 / 75.9 \\
& \textbf{Cosine-\textit{SimMark} (ours)} & 99.1 / 96.3 / 97.6 & \underline{98.9} / \underline{94.5} / \underline{96.4} & \underline{98.7} / \textbf{93.6} / \underline{96.1} & \underline{98.5} / \textbf{91.6} / \underline{96.0} & \underline{98.5} / \textbf{91.7} / \underline{95.5} & \underline{97.8} / \textbf{88.4} / \underline{94.7} & \underline{96.3} / \textbf{72.9} / \textbf{88.4}  & \underline{98.1} / \textbf{88.8} / \textbf{94.5} \\ 
& \textbf{Euclidean-\textit{SimMark}\footnotemark[7] (ours)} & \underline{99.8} / \underline{98.7} / 99.2 & \textbf{99.0} / \textbf{94.7} / \textbf{97.6} & \textbf{99.0} / \underline{91.9} / \textbf{96.2} & 97.8 / 75.9 / 89.5 & 97.7 / 76.4 / 90.4 & \textbf{98.7} / \underline{83.7} / \textbf{95.2} & \textbf{96.8} / \underline{65.8} / \underline{87.3}  & \textbf{98.2} / \underline{81.4} / \underline{92.7} \\ 
\bottomrule
\end{tabular} 
}
\caption{Performance of different algorithms across datasets and paraphrasers, evaluated using ROC-AUC $\uparrow$ / TP@FP=1\% $\uparrow$ / TP@FP=5\% $\uparrow$, respectively (↑: higher is better). In each column, \textbf{bold} value indicates the best performance for a given dataset and metric, while \underline{underlined} value denotes the second-best. \textit{SimMark} consistently outperforms or is on par with other state-of-the-art methods across datasets, paraphrasers, and is the best on average.}
\label{tab:performance}
\end{table*}

\begin{table}[t]
\small
\centering
\resizebox{0.38\textwidth}{!}{
\begin{tabular}{@{}lccc@{}}
\toprule
\textbf{Algorithm} & PPL $\downarrow$ & Ent-3 $\uparrow$ & Sem-Ent $\uparrow$ \\ \midrule
 No watermark       & 11.89                    & 11.43                    & 3.32                  
\\ \midrule
UW & 14.57 & 11.47 & 3.33 \\
 KGW                & 14.92                    & 11.32                    & 2.95                       \\
SIR                 & 20.34                    & 11.57                    & 3.18                       \\
SemStamp           & 12.89                   & 11.50                   & 3.32                      \\
$k$-SemStamp       & 11.82                    & 11.48                    & 3.32                      \\
\textbf{\textit{SimMark} (ours)} & 12.69                    & 11.50                    & 3.37                      \\
\bottomrule
\end{tabular}
}
\caption{Comparison of the quality of text watermarked using different algorithms on the BookSum dataset (↓: lower is better, ↑: higher is better). \textit{SimMark} yields quality metrics comparable to the no-watermark baseline, indicating minimal impact on text quality and semantic diversity. In contrast, token-level methods (UW, KGW, and SIR) notably degrade the text quality, especially in terms of perplexity.
}
\label{tab:quality}
\end{table}
\renewcommand*{\thefootnote}{\arabic{footnote}}
Performance of \textit{SimMark} is evaluated across different datasets and models using the area under the receiver operating characteristic curve (ROC-AUC) and true positive rate (TP) at fixed FP rates of 1\% and 5\% (TP@1\%FP and TP@5\%FP). Higher values indicate better performance across all metrics\footnote{Like \cite{hou-etal-2024-semstamp}, all results are from a single run.}. 

For dimensionality reduction, we fitted a PCA model on 8000 samples from the RealNews subset of the C4 dataset \cite{c4}, reducing embedding dimensions from 768 to 16. After testing various principal component counts (ranging from 512 to 16), we found 16 to yield the best results.  During our experiments, we evaluated both settings (with and without PCA). Specifically, PCA improved robustness against paraphrasing attacks when Euclidean distance was used (except on the BookSum dataset), but consistently degraded performance when cosine similarity was employed across all datasets. The results of these experiments are summarized in Table~\ref{tab:pca} in Appendix \ref{appendix:ablation_pca}.

Across all experiments, the decay factor was set to $K=250$, as this value provided an optimal trade-off between performance under both non-paraphrased and paraphrased conditions (see Appendix \ref{appendix:K} for an ablation study on this). The threshold $\beta$ was determined empirically during the detection (refer to Appendix \ref{appendix:estimate-AUC} for details) to achieve the specified FP rates (1\% or 5\%). The intervals $[0.68, 0.76]$ for cosine similarity and $[0.28, 0.36]$ for Euclidean distance with PCA and $[0.4, 0.55]$ for Euclidean distance without PCA were found to be near-optimal, as detailed in Appendix \ref{appendix:optimal_interval}. 

While the intervals as well as other hyperparameters such as decay factor $K$ could have been further optimized for each dataset or paraphraser individually, we chose not to do so to show the general performance of our method (in contrast, $k$-SemStamp fine-tunes domain-specific embedders that are optimized per setting).

\subsection{Models and Datasets}

For our experiments, we used the same fine-tuned version of OPT-1.3B \cite{zhang2022opt} as in \citet{hou-etal-2024-semstamp, hou-etal-2024-k}\footnote{Used \href{https://huggingface.co/AbeHou/opt-1.3b-semstamp}{AbeHou/opt-1.3b-semstamp} (1.3B) model.} to ensure fair comparison. However, we emphasize that our method is model-agnostic and it treats the LLM as a black-box text generator. As such, if the method performs well on one family of models, it is expected to generalize to others. To support this, we also tested our method on Gemma3-4B model \cite{gemma3} and observed similar results (see Appendix~\ref{appendix:gemma3}). For semantic embedding, we utilized Instructor-Large\footnote{Used \href{https://huggingface.co/hkunlp/instructor-large}{hkunlp/instructor-large} (335M) model.} \cite{su-etal-2023-one}. Appendix~\ref{appendix:eval-setting} includes additional details on the experimental configurations.

\renewcommand*{\thefootnote}{\fnsymbol{footnote}}
\footnotetext[7]{PCA is applied.}
\renewcommand*{\thefootnote}{\arabic{footnote}}
In our experiments, we used three English datasets: RealNews subset of C4\footnote{Dataset card: \href{https://huggingface.co/datasets/allenai/c4}{allenai/c4} (Validation split)} \cite{c4}, BookSum\footnote{Dataset card: \href{https://huggingface.co/datasets/kmfoda/booksum}{kmfoda/booksum} (Validation split)} \cite{kryscinski-etal-2022-booksum}, and Reddit-TIFU\footnote{Dataset card: \href{https://huggingface.co/datasets/ctr4si/reddit_tifu}{ctr4si/reddit\_tifu} (Train split, short subset)} \cite{kim-etal-2019-abstractive} datasets, as in \citet{hou-etal-2024-semstamp}. Specifically, 1000 samples from each dataset were chosen to analyze the detection performance and the text quality. Each sample was segmented into sentences\footnote{Using \textit{sent\_tokenize} method of NLTK \cite{bird2009natural}.}, with the first sentence serving as the \textit{prompt} to the LLM.

We evaluated text quality after applying \textit{SimMark} using the following metrics:
\begin{itemize}[noitemsep, topsep=0pt]
    \item \textbf{Tri-gram Entropy (Ent-3) ↑} \cite{ent-3}: Assesses textual diversity via the entropy of the tri-grams distribution.
    \item \textbf{Semantic Entropy (Sem-Ent) ↑} \cite{han-etal-2022-measuring}: Measures semantic informativeness and diversity of the text.
    \item \textbf{Perplexity (PPL) ↓} \cite{jelinek1977perplexity}: Measures how surprising the text is to an \textit{oracle} LLM\footnote{Used \href{https://huggingface.co/facebook/opt-2.7b}{facebook/opt-2.7b}, following \citet{hou-etal-2024-semstamp, hou-etal-2024-k}.}.
\end{itemize}
\subsection{Paraphrase Attack}
To evaluate the robustness of \textit{SimMark} against paraphrase attacks, we tested it using three paraphrasers: $I.$ Pegasus paraphraser\footnote{Used \href{https://huggingface.co/tuner007/pegasus_paraphrase}{tuner007/pegasus\_paraphrase} (568M) model.} \cite{pegasus}, $II.$ Parrot paraphraser\footnote{Used \href{https://huggingface.co/prithivida/parrot_paraphraser_on_T5}{parrot\_paraphraser\_on\_T5} (220M) model.} \cite{parrot}, $III.$ GPT-3.5-Turbo \cite{openai2022chatgpt}. Refer to Appendix \ref{appendix:prompt} to find the prompts used with GPT-3.5-Turbo. \citet{kirchenbauerreliability} observed that prompting models to paraphrase entire texts often results in summarized outputs, with the summarization ratio worsening for longer inputs. To prevent any information loss caused by such summarization, we adopted a sentence-by-sentence paraphrasing scheme, which also ensures our results are comparable to \citet{hou-etal-2024-semstamp, hou-etal-2024-k}. 
The quality of paraphrases was assessed using BertScore\footnote{Used \href{https://huggingface.co/microsoft/deberta-xlarge-mnli}{deberta-xlarge-mnli} (750M) \cite{he2021deberta}.} \cite{bertscore}, with all settings consistent with \citet{hou-etal-2024-semstamp, hou-etal-2024-k}. The bottom part of Figure~\ref{fig:Overview} demonstrates the paraphrase attack and detection phase in more detail.

We also included the results for the \textit{bigram} paraphrase attack introduced by \citet{hou-etal-2024-semstamp}, with identical settings (25 rephrases for each sentence when using Pegasus and Parrot, etc.). This attack involves generating multiple paraphrases for each sentence, and choosing the one that increases the likelihood of disrupting statistical signatures embedded in the text, especially for token-level algorithms \cite{hou-etal-2024-semstamp}. While this attack significantly impacts most other methods, \textit{SimMark} demonstrates greater robustness against it. We must highlight that $k$-SemStamp relies on domain-specific clustering of semantic spaces, making it domain-dependent. In contrast, both \textit{SimMark} and SemStamp are domain-independent. Still, \textit{SimMark} outperforms both in nearly all cases across various datasets and metrics, further underscoring its universality and robustness.  

\subsection{Robustness to Sentence-Level Perturbations}
To further evaluate the robustness of our method and its real-world applicability, we introduce more challenging attack scenarios: \textit{Paraphrase+Drop} and \textit{Paraphrase+Merge} Attacks. These scenarios simulate realistic adversarial editing strategies where a user attempts to remove or merge sentences after paraphrasing, while maintaining fluency.

\textit{Paraphrase+Drop} assumes that an adversary not only paraphrases the text but also drops sentences deemed redundant. This reflects common editing practices, especially since LLMs often produce verbose outputs due to imperfect reward modeling \cite{chiang2024over}. To simulate this, we first paraphrase the input text and then randomly drop sentences with a specified probability $p$. Similarly, \textit{Paraphrase+Merge} is designed to test robustness under more subtle structural changes. After paraphrasing, we replace end-of-sentence punctuations (e.g., \texttt{.}, \texttt{?}, \texttt{!}) with the word ``and'' with probability $0 < p < \frac{1}{2}$. In our experiments, we avoid higher values of $p$ as they result in unnaturally long and less fluent sentences. This setup simulates a realistic scenario where an adversary attempts to merge sentences while preserving the overall coherence of the text. These combined attacks serve to stress-test the resilience of watermarking methods under more naturalistic adversarial conditions that go beyond simple paraphrasing.

\subsection{Results \& Discussion}

\begin{figure}[t]
    \centering
    \includegraphics[width=0.95\linewidth]{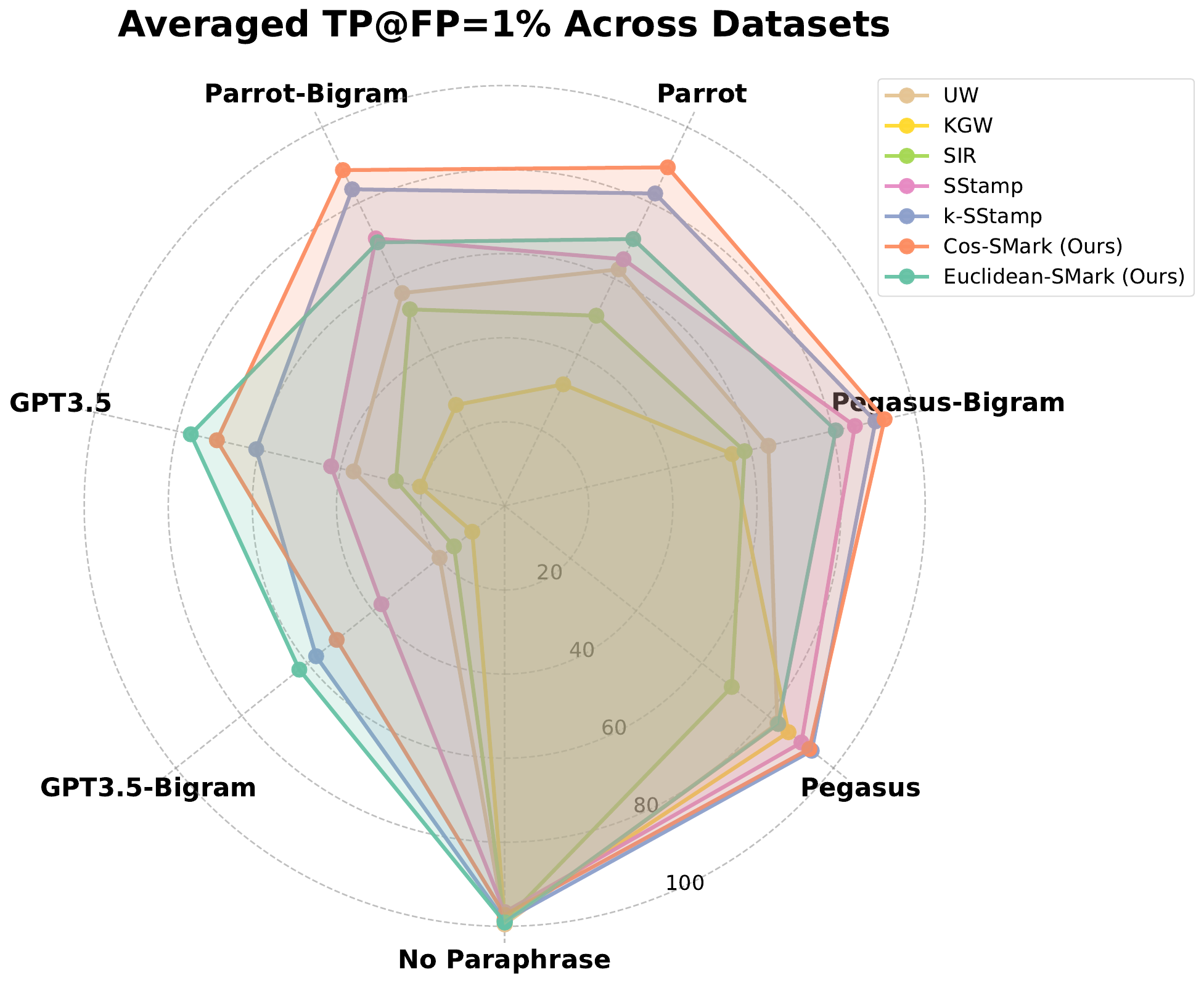}
    \caption{Detection performance of different watermarking methods under various paraphrasing attacks, measured by TP@FP=1\% ↑ and averaged across all three datasets (RealNews, BookSum, Reddit-TIFU). Each axis corresponds to a specific paraphrasing attack method (e.g., Pegasus-Bigram), and higher values are better. Our methods, \textcolor{Orange}{cosine-\textit{SimMark}} and \textcolor{Emerald}{Euclidean-\textit{SimMark}}, consistently outperform or match baselines across most paraphrasers, especially under more challenging conditions such as bigram-level paraphrasing.}
    \label{fig:radar1}
\end{figure}

\begin{figure}[t]
    \centering
    \includegraphics[width=0.48\textwidth]{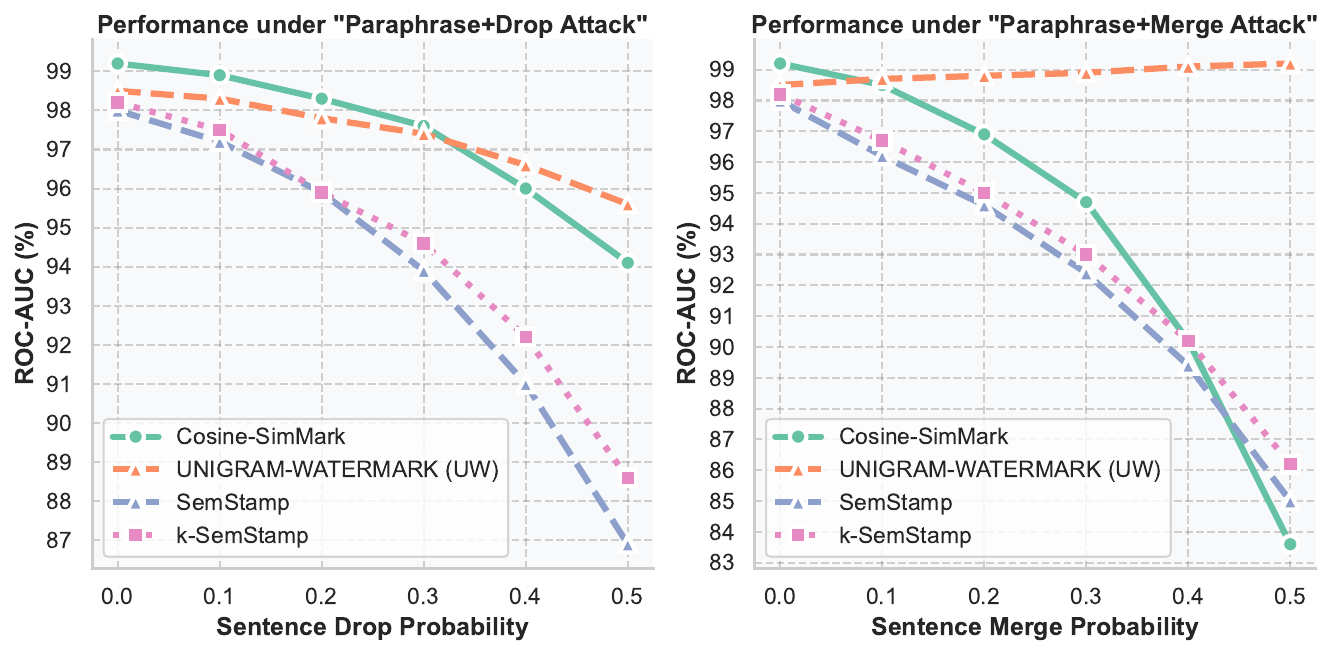}
        \caption{Detection performance (ROC-AUC ↑) under two adversarial settings using RealNews dataset. \textbf{Left}: \textit{Paraphrase+Drop Attack}, where random sentences are removed after paraphrasing. \textit{SimMark}, in nearly all parameter regimes, outperforms other methods under this attack. \textbf{Right}: \textit{Paraphrase+Merge Attack}, where sentence boundaries (punctuations) are probabilistically replaced with “and” to merge sentences. Although \textit{SimMark} performs best among sentence-level approaches, UW remains highly robust due to its token-level nature.}
    \label{fig:auc_plot}
\end{figure}

We compared the performance of \textit{SimMark} against SOTA watermarking algorithms through extensive experiments. Our primary baseline was SemStamp, a sentence-level semantic watermarking method. We also included $k$-SemStamp, an improvement over SemStamp tailored to specific domains. Results for SemStamp, $k$-SemStamp, KGW, and SIR, were extracted directly from \citet{hou-etal-2024-semstamp, hou-etal-2024-k}\footnote{Despite \citet{hou-etal-2024-semstamp, hou-etal-2024-k} releasing their code and data, we were unable to reproduce their reported results fully. Consequently, there are minor discrepancies between our reproduction results (shown in Figure~\ref{fig:auc_plot} for cases with $p=0$) and those presented in Table~\ref{tab:performance} (extracted directly from their paper). Additionally, the results reported in Table~\ref{tab:quality} (our reproduction) also show slight differences from their paper, likely due to hyperparameter details that were not explicitly documented.}.

Table~\ref{tab:performance} presents detection performance pre and post paraphrase attacks, while Table~\ref{tab:quality} provides text quality evaluation results. To better visualize robustness across paraphrasing attacks, we aggregate TP@FP=1\% scores from Table~\ref{tab:performance} across datasets and present them in a radar plot (Figure~\ref{fig:radar1}). Additional radar plots and a summary table comparing performance across all paraphrasing settings are provided in Appendix~\ref{appendix:average-performance} (see Table~\ref{tab:avg_performance_comparison} and Figure~\ref{fig:radar2}). We also conducted a small-scale A/B test to assess watermark imperceptibility. To see the results, please refer to Appendix \ref{appendix:human_eval}. Overall, our algorithm was imperceptible to the evaluators, and impacted the text quality minimally while being effective and consistently outperforming or matching other SOTA methods, achieving the highest average paraphrased performance across all datasets. 

Notably, our method, \textit{SimMark} (domain-independent), surpasses the primary baseline, SemStamp (domain-independent), and is on par with or exceeds $k$-SemStamp (domain-dependent). A key aspect to consider is that the fine-tuning of SemStamp and $k$-SemStamp’s embedding model on text paraphrased by Pegasus likely contributes to their higher robustness against Pegasus but may introduce bias and reduce general applicability (as shown in Table 2 of \citet{hou-etal-2024-k}, $k$-SemStamp loses performance under domain shift). Additionally, the results for the Reddit-TIFU dataset were only available for SemStamp and not $k$-SemStamp, likely due to the dataset’s informal, diverse text style and $k$-SemStamp's limitation for text to belong to a specific domain, such as news articles or scientific writings~\cite{{hou-etal-2024-k}}.

While \textit{SimMark} demonstrates strong performance, we acknowledge that it is not always the best across every setting. \textit{SimMark} is designed with generality, robustness, and black-box compatibility in mind. It forgoes domain-specific fine-tuning in favor of broader applicability, which may explain why it does not always outperform more narrowly optimized methods. Several factors may explain its relative underperformance in some cases:

First, some variability arises from randomness in datasets and generation. LLMs rely on pseudo-random decoding, which may introduce subtle fluctuations in outputs. More importantly, as \textit{SimMark} operates at the sentence level, it cannot fine-tune token-level patterns similar to token-level methods (e.g., UW), which may subtly manipulate every token to embed statistical signals. This granularity difference can be particularly advantageous when generation length is short (e.g., a 200-token cap). In such cases, token-level methods benefit from having more embedding capacities, whereas sentence-level methods like \textit{SimMark} may yield weaker statistical signals, resulting in greater performance variability in shorter texts. In other words, token-level methods benefit from high-frequency watermark signals, whereas \textit{SimMark} injects at a coarser granularity. As a result, for \textit{SimMark}, detection performance improves with longer generations, where more sentence-level signal can accumulate.

Figure~\ref{fig:auc_plot} presents the ROC-AUC performance of UW (the best token-level method in our experiments), \textit{SimMark}, $k$-SemStamp, and SemStamp under \textit{Paraphrase+Drop} and \textit{Paraphrase+Merge} attacks, evaluated on the RealNews dataset. Under \textit{Paraphrase+Drop}, across most parameter regimes, \textit{SimMark} outperforms all other methods, sustaining higher detection performance even as the attack intensity increases. While \textit{SimMark} shows superior performance among sentence-based methods under \textit{Paraphrase+Merge}, UW maintains the highest robustness because it operates at the token level. 

Regarding sampling efficiency, for BookSum dataset for instance, \textit{SimMark} required an average of 7.1 samples per sentence from the LLM, compared to $k$-SemStamp and SemStamp, which averaged 13.3 and 20.9 samples, respectively \cite{{hou-etal-2024-k}}. This demonstrates that our method not only outperforms these baselines but is also 2-3 times more efficient (see Appendix \ref{appendix:sampling-efficiency} for theoretical estimates of this). For a comparison between \textit{SimMark} and token-level methods in terms of real-world runtime, see Appendix \ref{appendix:generation-efficiency}. Finally, refer to Appendix \ref{appendix:examples} for qualitative examples of \textit{SimMark}. 

\section{Conclusion} \label{conclusion}
In this paper, we introduced {\textit{SimMark}}, a similarity-based, robust sentence-level watermarking algorithm. Unlike existing approaches, \textit{SimMark} operates without requiring access to the internals of the model, ensuring compatibility with a wide range of LLMs, including API-only models. By utilizing a pre-trained general-purpose embedding model and integrating a \textit{soft} counting mechanism, \textit{SimMark} combines robustness against paraphrasing with applicability to diverse domains. Experimental results show that \textit{SimMark} outperforms SOTA sentence-level watermarking algorithms in both efficiency and robustness to paraphrasing, representing a step forward in fully semantic watermarking for LLMs.

\section*{Limitations}
While \textit{SimMark} demonstrates outstanding performance, there are still some areas that warrant further exploration:

\paragraph{Rejection Sampling Overhead.} The rejection sampling process requires generating multiple candidate sentences until a valid sentence is accepted. Although our method is significantly (2-3 times) more efficient than prior approaches such as SemStamp and $k$-SemStamp, there is still a notable decrease in generation speed due to rejection sampling. Techniques like batch sampling or parallel sampling could potentially mitigate this issue, though at the expense of higher computational resource usage. Future research should focus on optimizing the method to balance efficiency and resource requirements.

\paragraph{Resistance to More Advanced Attacks.} While \textit{SimMark} demonstrates robustness against paraphrasing attacks, it may not be immune to more sophisticated adversarial transformations. In particular, detection could become less effective when watermarked text is interleaved with or embedded within a larger body of unwatermarked content. Additionally, although reverse engineering the exact watermarking rules is non-trivial, an adversary may attempt a spoofing attack by approximating our setup—for instance, by employing a publicly available embedding model or fitting a PCA model with publicly available datasets. While such attempts may not perfectly replicate the original embedding distribution, they could still pose a threat. We leave a thorough investigation into vulnerabilities and corresponding defences against reverse engineering to future work.

\paragraph{Dependency on Predefined Intervals.} In our experiments, we used consistent, predefined intervals across all datasets and observed consistently strong performance. Notably, we did not observe any noticeable degradation in text quality due to this interval constraint during rejection sampling (as shown in Table~\ref{tab:quality}), likely because the constraint applies only to consecutive sentences. Nonetheless, slight variations in the embeddings similarity distribution of LLM-generated text across different models/datasets may impact watermarking effectiveness. Adaptive strategies for setting these intervals dynamically (or pseudo-randomly) could not only improve performance but also make reverse-engineering the algorithm more difficult.

\section*{Ethical Considerations\footnote{Generative AI tools, such as ChatGPT, were used to refine this manuscript. The authors retain full responsibility for all content presented in the paper, ensuring adherence to academic integrity and ethical research standards.}}
\paragraph{Potential Risks.} By enabling robust detection of LLM-generated text, particularly under paraphrasing attacks, \textit{SimMark} tries to address ethical concerns surrounding the transparency and accountability of AI-generated content. However, like any watermarking algorithm, there are potential risks, such as falsely implicating human authors or adversaries developing more advanced techniques for spoofing attacks or bypassing detection. 
We acknowledge these limitations and advocate for the responsible deployment of such tools in combination with other verification mechanisms to mitigate these risks and ensure ethical, fair deployment. The primary goal of this work is to advance research in watermarking techniques to support the responsible use of LLMs. We believe that the societal impacts and ethical considerations of our work align with those outlined in \citet{weidinger2021ethical}.

\paragraph{Use of Models and Datasets.} Our research used datasets and pretrained models from the Hugging Face Hub\footnote{\url{https://huggingface.co}}, a public platform hosting machine learning resources under various licenses. We adhered to all license terms and intended usage guidelines for each artifact, which are documented on their individual Hugging Face model or dataset cards cited in the paper. All resources were used solely for research purposes in accordance with their intended use and respective licenses. The datasets employed are publicly available, widely used in prior research, and, to the best of our knowledge, free of personally identifiable information or offensive content. Any outputs generated by our method are intended for academic use, not for real-world or commercial applications, and no personal or sensitive data was processed. Any new artifacts we create (e.g., watermarked samples) are intended solely for academic evaluation, and we do not release any derivative data that violates original licensing terms.

\section*{Acknowledgments}
This work was supported by the NSERC Discovery Grant No. RGPIN-2019-05448.
\bibliography{anthology,acl_latex}

\begin{thebibliography}{50}
\providecommand{\natexlab}[1]{#1}

\bibitem[{Aaronson and Kirchner(2022)}]{aaronson2022}
Scott Aaronson and Hendrik Kirchner. 2022.
\newblock \href {https://www.scottaaronson.com/talks/watermark.ppt} {Watermarking gpt outputs}.

\bibitem[{Atallah et~al.(2001)Atallah, Raskin, Crogan, Hempelmann, Kerschbaum, Mohamed, and Naik}]{atallah2001natural}
Mikhail~J Atallah, Victor Raskin, Michael Crogan, Christian Hempelmann, Florian Kerschbaum, Dina Mohamed, and Sanket Naik. 2001.
\newblock Natural language watermarking: Design, analysis, and a proof-of-concept implementation.
\newblock In \emph{Information Hiding: 4th International Workshop, IH 2001 Pittsburgh, PA, USA, April 25--27, 2001 Proceedings 4}, pages 185--200. Springer.

\bibitem[{Bird et~al.(2009)Bird, Klein, and Loper}]{bird2009natural}
Steven Bird, Ewan Klein, and Edward Loper. 2009.
\newblock \emph{Natural language processing with Python: analyzing text with the natural language toolkit}.
\newblock " O'Reilly Media, Inc.".

\bibitem[{Chang et~al.(2024)Chang, Krishna, Houmansadr, Wieting, and Iyyer}]{chang-etal-2024-postmark}
Yapei Chang, Kalpesh Krishna, Amir Houmansadr, John~Frederick Wieting, and Mohit Iyyer. 2024.
\newblock \href {https://doi.org/10.18653/v1/2024.emnlp-main.506} {{P}ost{M}ark: A robust blackbox watermark for large language models}.
\newblock In \emph{Proceedings of the 2024 Conference on Empirical Methods in Natural Language Processing}, pages 8969--8987, Miami, Florida, USA. Association for Computational Linguistics.

\bibitem[{Chiang and Lee(2024)}]{chiang2024over}
Cheng-Han Chiang and Hung-yi Lee. 2024.
\newblock Over-reasoning and redundant calculation of large language models.
\newblock \emph{arXiv preprint arXiv:2401.11467}.

\bibitem[{Damodaran(2021)}]{parrot}
Prithiviraj Damodaran. 2021.
\newblock Parrot: Paraphrase generation for nlu.

\bibitem[{Fernandez et~al.(2023)Fernandez, Couairon, J\'egou, Douze, and Furon}]{Fernandez_2023_ICCV}
Pierre Fernandez, Guillaume Couairon, Herv\'e J\'egou, Matthijs Douze, and Teddy Furon. 2023.
\newblock The stable signature: Rooting watermarks in latent diffusion models.
\newblock In \emph{Proceedings of the IEEE/CVF International Conference on Computer Vision (ICCV)}, pages 22466--22477.

\bibitem[{Finlayson et~al.(2024)Finlayson, Ren, and Swayamdipta}]{finlayson2024logits}
Matthew Finlayson, Xiang Ren, and Swabha Swayamdipta. 2024.
\newblock Logits of api-protected llms leak proprietary information.
\newblock \emph{arXiv preprint arXiv:2403.09539}.

\bibitem[{Fu et~al.(2024)Fu, Xiong, and Dong}]{fu2024watermarking}
Yu~Fu, Deyi Xiong, and Yue Dong. 2024.
\newblock \href {https://doi.org/10.1609/aaai.v38i16.29756} {Watermarking conditional text generation for ai detection: Unveiling challenges and a semantic-aware watermark remedy}.
\newblock \emph{Proceedings of the AAAI Conference on Artificial Intelligence}, 38(16):18003--18011.

\bibitem[{Futurism(2023)}]{futurism2023cnet}
Futurism. 2023.
\newblock \href {https://futurism.com/cnet-ai-articles-label} {Cnet quietly deletes ai-generated articles amid backlash}.
\newblock Accessed: January 28, 2025.

\bibitem[{Hadsell et~al.(2006)Hadsell, Chopra, and LeCun}]{1640964}
R.~Hadsell, S.~Chopra, and Y.~LeCun. 2006.
\newblock \href {https://doi.org/10.1109/CVPR.2006.100} {Dimensionality reduction by learning an invariant mapping}.
\newblock In \emph{2006 IEEE Computer Society Conference on Computer Vision and Pattern Recognition (CVPR'06)}, volume~2, pages 1735--1742.

\bibitem[{Han et~al.(2022)Han, Kim, and Chang}]{han-etal-2022-measuring}
Seungju Han, Beomsu Kim, and Buru Chang. 2022.
\newblock \href {https://doi.org/10.18653/v1/2022.findings-emnlp.66} {Measuring and improving semantic diversity of dialogue generation}.
\newblock In \emph{Findings of the Association for Computational Linguistics: EMNLP 2022}, pages 934--950, Abu Dhabi, United Arab Emirates. Association for Computational Linguistics.

\bibitem[{Hao et~al.(2025)Hao, Qiang, Zhu, Li, Yuan, and Ouyang}]{hao-etal-2025-post}
Jifei Hao, Jipeng Qiang, Yi~Zhu, Yun Li, Yunhao Yuan, and Xiaoye Ouyang. 2025.
\newblock \href {https://aclanthology.org/2025.coling-main.364/} {Post-hoc watermarking for robust detection in text generated by large language models}.
\newblock In \emph{Proceedings of the 31st International Conference on Computational Linguistics}, pages 5430--5442, Abu Dhabi, UAE. Association for Computational Linguistics.

\bibitem[{He et~al.(2021)He, Liu, Gao, and Chen}]{he2021deberta}
Pengcheng He, Xiaodong Liu, Jianfeng Gao, and Weizhu Chen. 2021.
\newblock \href {https://openreview.net/forum?id=XPZIaotutsD} {Deberta: Decoding-enhanced bert with disentangled attention}.
\newblock In \emph{International Conference on Learning Representations}.

\bibitem[{Hou et~al.(2024{\natexlab{a}})Hou, Zhang, He, Wang, Chuang, Wang, Shen, Van~Durme, Khashabi, and Tsvetkov}]{hou-etal-2024-semstamp}
Abe Hou, Jingyu Zhang, Tianxing He, Yichen Wang, Yung-Sung Chuang, Hongwei Wang, Lingfeng Shen, Benjamin Van~Durme, Daniel Khashabi, and Yulia Tsvetkov. 2024{\natexlab{a}}.
\newblock \href {https://doi.org/10.18653/v1/2024.naacl-long.226} {{S}em{S}tamp: A semantic watermark with paraphrastic robustness for text generation}.
\newblock In \emph{Proceedings of the 2024 Conference of the North American Chapter of the Association for Computational Linguistics: Human Language Technologies (Volume 1: Long Papers)}, pages 4067--4082, Mexico City, Mexico. Association for Computational Linguistics.

\bibitem[{Hou et~al.(2024{\natexlab{b}})Hou, Zhang, Wang, Khashabi, and He}]{hou-etal-2024-k}
Abe Hou, Jingyu Zhang, Yichen Wang, Daniel Khashabi, and Tianxing He. 2024{\natexlab{b}}.
\newblock \href {https://doi.org/10.18653/v1/2024.findings-acl.98} {k-{S}em{S}tamp: A clustering-based semantic watermark for detection of machine-generated text}.
\newblock In \emph{Findings of the Association for Computational Linguistics: ACL 2024}, pages 1706--1715, Bangkok, Thailand. Association for Computational Linguistics.

\bibitem[{Hurst et~al.(2024)Hurst, Lerer, Goucher, Perelman, Ramesh, Clark, Ostrow, Welihinda, Hayes, Radford et~al.}]{hurst2024gpt}
Aaron Hurst, Adam Lerer, Adam~P Goucher, Adam Perelman, Aditya Ramesh, Aidan Clark, AJ~Ostrow, Akila Welihinda, Alan Hayes, Alec Radford, and 1 others. 2024.
\newblock Gpt-4o system card.
\newblock \emph{arXiv preprint arXiv:2410.21276}.

\bibitem[{Indyk and Motwani(1998)}]{lsh}
Piotr Indyk and Rajeev Motwani. 1998.
\newblock Approximate nearest neighbors: towards removing the curse of dimensionality.
\newblock In \emph{Proceedings of the thirtieth annual ACM symposium on Theory of computing}, pages 604--613.

\bibitem[{Jelinek et~al.(1977)Jelinek, Mercer, Bahl, and Baker}]{jelinek1977perplexity}
Fred Jelinek, Robert~L Mercer, Lalit~R Bahl, and James~K Baker. 1977.
\newblock Perplexity—a measure of the difficulty of speech recognition tasks.
\newblock \emph{The Journal of the Acoustical Society of America}, 62(S1):S63--S63.

\bibitem[{Jolliffe(2002)}]{jolliffe2002principal}
Ian~T Jolliffe. 2002.
\newblock \emph{Principal component analysis for special types of data}.
\newblock Springer.

\bibitem[{Kim et~al.(2019)Kim, Kim, and Kim}]{kim-etal-2019-abstractive}
Byeongchang Kim, Hyunwoo Kim, and Gunhee Kim. 2019.
\newblock \href {https://doi.org/10.18653/v1/N19-1260} {Abstractive summarization of {R}eddit posts with multi-level memory networks}.
\newblock In \emph{Proceedings of the 2019 Conference of the North {A}merican Chapter of the Association for Computational Linguistics: Human Language Technologies, Volume 1 (Long and Short Papers)}, pages 2519--2531, Minneapolis, Minnesota. Association for Computational Linguistics.

\bibitem[{Kirchenbauer et~al.(2023)Kirchenbauer, Geiping, Wen, Katz, Miers, and Goldstein}]{kgw}
John Kirchenbauer, Jonas Geiping, Yuxin Wen, Jonathan Katz, Ian Miers, and Tom Goldstein. 2023.
\newblock A watermark for large language models.
\newblock In \emph{International Conference on Machine Learning}, pages 17061--17084. PMLR.

\bibitem[{Kirchenbauer et~al.(2024)Kirchenbauer, Geiping, Wen, Shu, Saifullah, Kong, Fernando, Saha, Goldblum, and Goldstein}]{kirchenbauerreliability}
John Kirchenbauer, Jonas Geiping, Yuxin Wen, Manli Shu, Khalid Saifullah, Kezhi Kong, Kasun Fernando, Aniruddha Saha, Micah Goldblum, and Tom Goldstein. 2024.
\newblock \href {https://openreview.net/forum?id=DEJIDCmWOz} {On the reliability of watermarks for large language models}.
\newblock In \emph{The Twelfth International Conference on Learning Representations}.

\bibitem[{Krishna et~al.(2024)Krishna, Song, Karpinska, Wieting, and Iyyer}]{krishna2024paraphrasing}
Kalpesh Krishna, Yixiao Song, Marzena Karpinska, John Wieting, and Mohit Iyyer. 2024.
\newblock Paraphrasing evades detectors of ai-generated text, but retrieval is an effective defense.
\newblock \emph{Advances in Neural Information Processing Systems}, 36.

\bibitem[{Kryscinski et~al.(2022)Kryscinski, Rajani, Agarwal, Xiong, and Radev}]{kryscinski-etal-2022-booksum}
Wojciech Kryscinski, Nazneen Rajani, Divyansh Agarwal, Caiming Xiong, and Dragomir Radev. 2022.
\newblock \href {https://doi.org/10.18653/v1/2022.findings-emnlp.488} {{BOOKSUM}: A collection of datasets for long-form narrative summarization}.
\newblock In \emph{Findings of the Association for Computational Linguistics: EMNLP 2022}, pages 6536--6558, Abu Dhabi, United Arab Emirates. Association for Computational Linguistics.

\bibitem[{Kumarage et~al.(2023)Kumarage, Sheth, Moraffah, Garland, and Liu}]{kumarage-etal-2023-reliable}
Tharindu Kumarage, Paras Sheth, Raha Moraffah, Joshua Garland, and Huan Liu. 2023.
\newblock \href {https://doi.org/10.18653/v1/2023.findings-emnlp.94} {How reliable are {AI}-generated-text detectors? an assessment framework using evasive soft prompts}.
\newblock In \emph{Findings of the Association for Computational Linguistics: EMNLP 2023}, pages 1337--1349, Singapore. Association for Computational Linguistics.

\bibitem[{Kwon et~al.(2023)Kwon, Li, Zhuang, Sheng, Zheng, Yu, Gonzalez, Zhang, and Stoica}]{kwon2023efficient}
Woosuk Kwon, Zhuohan Li, Siyuan Zhuang, Ying Sheng, Lianmin Zheng, Cody~Hao Yu, Joseph~E. Gonzalez, Hao Zhang, and Ion Stoica. 2023.
\newblock Efficient memory management for large language model serving with pagedattention.
\newblock In \emph{Proceedings of the ACM SIGOPS 29th Symposium on Operating Systems Principles}.

\bibitem[{Liu et~al.(2023)Liu, Pan, Hu, Meng, and Wen}]{sir}
Aiwei Liu, Leyi Pan, Xuming Hu, Shiao Meng, and Lijie Wen. 2023.
\newblock A semantic invariant robust watermark for large language models.
\newblock \emph{arXiv preprint arXiv:2310.06356}.

\bibitem[{Lloyd(1982)}]{kmeans}
Stuart Lloyd. 1982.
\newblock Least squares quantization in pcm.
\newblock \emph{IEEE transactions on information theory}, 28(2):129--137.

\bibitem[{OpenAI(2022)}]{openai2022chatgpt}
OpenAI. 2022.
\newblock \href {https://openai.com/blog/chatgpt} {{ChatGPT}}.

\bibitem[{Pan et~al.(2024)Pan, Liu, He, Gao, Zhao, Lu, Zhou, Liu, Hu, Wen, King, and Yu}]{pan-etal-2024-markllm}
Leyi Pan, Aiwei Liu, Zhiwei He, Zitian Gao, Xuandong Zhao, Yijian Lu, Binglin Zhou, Shuliang Liu, Xuming Hu, Lijie Wen, Irwin King, and Philip~S. Yu. 2024.
\newblock \href {https://doi.org/10.18653/v1/2024.emnlp-demo.7} {{M}ark{LLM}: An open-source toolkit for {LLM} watermarking}.
\newblock In \emph{Proceedings of the 2024 Conference on Empirical Methods in Natural Language Processing: System Demonstrations}, pages 61--71, Miami, Florida, USA. Association for Computational Linguistics.

\bibitem[{Raffel et~al.(2020)Raffel, Shazeer, Roberts, Lee, Narang, Matena, Zhou, Li, and Liu}]{c4}
Colin Raffel, Noam Shazeer, Adam Roberts, Katherine Lee, Sharan Narang, Michael Matena, Yanqi Zhou, Wei Li, and Peter~J. Liu. 2020.
\newblock \href {http://jmlr.org/papers/v21/20-074.html} {Exploring the limits of transfer learning with a unified text-to-text transformer}.
\newblock \emph{Journal of Machine Learning Research}, 21(140):1--67.

\bibitem[{Sadasivan et~al.(2023)Sadasivan, Kumar, Balasubramanian, Wang, and Feizi}]{reliable}
Vinu~Sankar Sadasivan, Aounon Kumar, Sriram Balasubramanian, Wenxiao Wang, and Soheil Feizi. 2023.
\newblock Can ai-generated text be reliably detected?
\newblock \emph{arXiv preprint arXiv:2303.11156}.

\bibitem[{Singhal et~al.(2023)Singhal, Azizi, Tu, Mahdavi, Wei, Chung, Scales, Tanwani, Cole-Lewis, Pfohl et~al.}]{singhal2023large}
Karan Singhal, Shekoofeh Azizi, Tao Tu, S~Sara Mahdavi, Jason Wei, Hyung~Won Chung, Nathan Scales, Ajay Tanwani, Heather Cole-Lewis, Stephen Pfohl, and 1 others. 2023.
\newblock Large language models encode clinical knowledge.
\newblock \emph{Nature}, 620(7972):172--180.

\bibitem[{Su et~al.(2023)Su, Shi, Kasai, Wang, Hu, Ostendorf, Yih, Smith, Zettlemoyer, and Yu}]{su-etal-2023-one}
Hongjin Su, Weijia Shi, Jungo Kasai, Yizhong Wang, Yushi Hu, Mari Ostendorf, Wen-tau Yih, Noah~A. Smith, Luke Zettlemoyer, and Tao Yu. 2023.
\newblock \href {https://doi.org/10.18653/v1/2023.findings-acl.71} {One embedder, any task: Instruction-finetuned text embeddings}.
\newblock In \emph{Findings of the Association for Computational Linguistics: ACL 2023}, pages 1102--1121, Toronto, Canada. Association for Computational Linguistics.

\bibitem[{Taranukhin et~al.(2024)Taranukhin, Ravi, Lukacs, Milios, and Shwartz}]{taranukhin-etal-2024-empowering}
Maksym Taranukhin, Sahithya Ravi, Gabor Lukacs, Evangelos Milios, and Vered Shwartz. 2024.
\newblock \href {https://doi.org/10.18653/v1/2024.nllp-1.27} {Empowering air travelers: A chatbot for {C}anadian air passenger rights}.
\newblock In \emph{Proceedings of the Natural Legal Language Processing Workshop 2024}, pages 326--335, Miami, FL, USA. Association for Computational Linguistics.

\bibitem[{Team et~al.(2025)Team, Kamath, Ferret, Pathak, Vieillard, Merhej, Perrin, Matejovicova, Ram{\'e}, Rivi{\`e}re et~al.}]{gemma3}
Gemma Team, Aishwarya Kamath, Johan Ferret, Shreya Pathak, Nino Vieillard, Ramona Merhej, Sarah Perrin, Tatiana Matejovicova, Alexandre Ram{\'e}, Morgane Rivi{\`e}re, and 1 others. 2025.
\newblock Gemma 3 technical report.
\newblock \emph{arXiv preprint arXiv:2503.19786}.

\bibitem[{Teymoorianfard et~al.(2025)Teymoorianfard, Ma, and Houmansadr}]{teymoorianfard2025vidstamptemporallyawarewatermarkownership}
Mohammadreza Teymoorianfard, Shiqing Ma, and Amir Houmansadr. 2025.
\newblock \href {https://arxiv.org/abs/2505.01406} {Vidstamp: A temporally-aware watermark for ownership and integrity in video diffusion models}.
\newblock \emph{Preprint}, arXiv:2505.01406.

\bibitem[{Topkara et~al.(2006)Topkara, Topkara, and Atallah}]{topkara2006words}
Mercan Topkara, Umut Topkara, and Mikhail~J Atallah. 2006.
\newblock Words are not enough: sentence level natural language watermarking.
\newblock In \emph{Proceedings of the 4th ACM international workshop on Contents protection and security}, pages 37--46.

\bibitem[{Torabi et~al.(2025)Torabi, Shirani, and Reilly}]{torabi2025large}
Yasaman Torabi, Shahram Shirani, and James~P Reilly. 2025.
\newblock Large language model-based nonnegative matrix factorization for cardiorespiratory sound separation.
\newblock \emph{arXiv preprint arXiv:2502.05757}.

\bibitem[{Weidinger et~al.(2021)Weidinger, Mellor, Rauh, Griffin, Uesato, Huang, Cheng, Glaese, Balle, Kasirzadeh et~al.}]{weidinger2021ethical}
Laura Weidinger, John Mellor, Maribeth Rauh, Conor Griffin, Jonathan Uesato, Po-Sen Huang, Myra Cheng, Mia Glaese, Borja Balle, Atoosa Kasirzadeh, and 1 others. 2021.
\newblock Ethical and social risks of harm from language models.
\newblock \emph{arXiv preprint arXiv:2112.04359}.

\bibitem[{Wolf et~al.(2020)Wolf, Debut, Sanh, Chaumond, Delangue, Moi, Cistac, Rault, Louf, Funtowicz, Davison, Shleifer, von Platen, Ma, Jernite, Plu, Xu, Le~Scao, Gugger, Drame, Lhoest, and Rush}]{wolf-etal-2020-transformers}
Thomas Wolf, Lysandre Debut, Victor Sanh, Julien Chaumond, Clement Delangue, Anthony Moi, Pierric Cistac, Tim Rault, Remi Louf, Morgan Funtowicz, Joe Davison, Sam Shleifer, Patrick von Platen, Clara Ma, Yacine Jernite, Julien Plu, Canwen Xu, Teven Le~Scao, Sylvain Gugger, and 3 others. 2020.
\newblock \href {https://doi.org/10.18653/v1/2020.emnlp-demos.6} {Transformers: State-of-the-art natural language processing}.
\newblock In \emph{Proceedings of the 2020 Conference on Empirical Methods in Natural Language Processing: System Demonstrations}, pages 38--45, Online. Association for Computational Linguistics.

\bibitem[{Wu et~al.(2025)Wu, Deng, Li, Liu, Mi, Peng, Xu, Liu, Cho, Choi et~al.}]{wu2025medreason}
Juncheng Wu, Wenlong Deng, Xingxuan Li, Sheng Liu, Taomian Mi, Yifan Peng, Ziyang Xu, Yi~Liu, Hyunjin Cho, Chang-In Choi, and 1 others. 2025.
\newblock Medreason: Eliciting factual medical reasoning steps in llms via knowledge graphs.
\newblock \emph{arXiv preprint arXiv:2504.00993}.

\bibitem[{Yang et~al.(2023)Yang, Chen, Zhang, Liu, Qi, Zhang, Fang, and Yu}]{yang2023watermarking}
Xi~Yang, Kejiang Chen, Weiming Zhang, Chang Liu, Yuang Qi, Jie Zhang, Han Fang, and Nenghai Yu. 2023.
\newblock Watermarking text generated by black-box language models.
\newblock \emph{arXiv preprint arXiv:2305.08883}.

\bibitem[{Yao et~al.(2024)Yao, Ke, Wang, Li, and Hu}]{yao2024lawyer}
Shunyu Yao, Qingqing Ke, Qiwei Wang, Kangtong Li, and Jie Hu. 2024.
\newblock Lawyer gpt: A legal large language model with enhanced domain knowledge and reasoning capabilities.
\newblock In \emph{Proceedings of the 2024 3rd International Symposium on Robotics, Artificial Intelligence and Information Engineering}, pages 108--112.

\bibitem[{Zhang et~al.(2020)Zhang, Zhao, Saleh, and Liu}]{pegasus}
Jingqing Zhang, Yao Zhao, Mohammad Saleh, and Peter Liu. 2020.
\newblock \href {https://proceedings.mlr.press/v119/zhang20ae.html} {{PEGASUS}: Pre-training with extracted gap-sentences for abstractive summarization}.
\newblock In \emph{Proceedings of the 37th International Conference on Machine Learning}, volume 119 of \emph{Proceedings of Machine Learning Research}, pages 11328--11339. PMLR.

\bibitem[{Zhang et~al.(2022)Zhang, Roller, Goyal, Artetxe, Chen, Chen, Dewan, Diab, Li, Lin et~al.}]{zhang2022opt}
Susan Zhang, Stephen Roller, Naman Goyal, Mikel Artetxe, Moya Chen, Shuohui Chen, Christopher Dewan, Mona Diab, Xian Li, Xi~Victoria Lin, and 1 others. 2022.
\newblock Opt: Open pre-trained transformer language models.
\newblock \emph{arXiv preprint arXiv:2205.01068}.

\bibitem[{Zhang* et~al.(2020)Zhang*, Kishore*, Wu*, Weinberger, and Artzi}]{bertscore}
Tianyi Zhang*, Varsha Kishore*, Felix Wu*, Kilian~Q. Weinberger, and Yoav Artzi. 2020.
\newblock \href {https://openreview.net/forum?id=SkeHuCVFDr} {Bertscore: Evaluating text generation with bert}.
\newblock In \emph{International Conference on Learning Representations}.

\bibitem[{Zhang et~al.(2018)Zhang, Galley, Gao, Gan, Li, Brockett, and Dolan}]{ent-3}
Yizhe Zhang, Michel Galley, Jianfeng Gao, Zhe Gan, Xiujun Li, Chris Brockett, and Bill Dolan. 2018.
\newblock Generating informative and diverse conversational responses via adversarial information maximization.
\newblock \emph{Advances in Neural Information Processing Systems}, 31.

\bibitem[{Zhao et~al.(2023)Zhao, Ananth, Li, and Wang}]{unigram}
Xuandong Zhao, Prabhanjan Ananth, Lei Li, and Yu-Xiang Wang. 2023.
\newblock Provable robust watermarking for ai-generated text.
\newblock \emph{arXiv preprint arXiv:2306.17439}.

\end{thebibliography}
\newpage
\appendix
\setcounter{footnote}{0}
\section*{Supplemental Materials}
\section{Aditional Related Work}
\label{appendix:additional-related}
\subsection{Token-Level Watermarking}
\citeposs{unigram} UNIGRAM-WATERMARK (UW) builds upon KGW by fixing the red and green lists instead of pseudo-randomly selecting them, proving that, compared to KGW, their method is more robust to paraphrasing and editing \cite{unigram}. However, as outlined by \citet{{hou-etal-2024-semstamp}}, this algorithm can be reverse-engineered, rendering it impractical for high-stakes, real-world applications (for details about the reverse-engineering procedure, refer to \citet{{hou-etal-2024-semstamp}}).

The Semantic Invariant Robust (SIR) watermark in \citet{sir} is also similar to KGW but is designed to be less sensitive to attacks involving synonym replacement or advanced paraphrasing. SIR achieves this by altering the LLM logits based on the semantics of previously generated tokens, using a semantic embedding model to compute semantic representations, and training a model that adjusts LLM's logits based on the semantic embeddings of prior tokens \cite{sir}.  
\subsection{Post-Hoc Watermarking}

\citeposs{chang-etal-2024-postmark} PostMark is a post-hoc watermarking algorithm designed to work without access to model logits, making it compatible with API-only LLMs. It constructs an input-dependent set of candidate words using semantic embeddings and then prompts another LLM (e.g., GPT-4o) to insert these words into the generated text. Detection relies on statistical analysis of the inserted words. While PostMark’s compatibility with black-box LLMs is a strength, the approach is expensive—watermarking 100 tokens is estimated to cost around \$1.2 USD \cite{chang-etal-2024-postmark}.

\citet{yang2023watermarking} propose another post-hoc method that encodes each word in the text as a binary bit via a Bernoulli distribution ($p=0.5$), embedding the watermark through synonym substitution: words representing bit 0 are replaced with synonyms representing bit 1. Detection is again done via statistical testing. However, this method is fragile: synonyms are not reliably preserved under paraphrasing and often fail to capture subtle contextual meanings, which can noticeably degrade text quality and watermark robustness.

\citet{hao-etal-2025-post} is a post-hoc watermarking technique similar to \citet{yang2023watermarking} that improves robustness by selecting semantically or syntactically essential words—those less likely to be altered during paraphrasing—as anchor points for embedding. The method uses paraphrase-based lexical substitution to insert watermarks while preserving the original semantics. However, empirical results in \citet{chang-etal-2024-postmark} demonstrate that this method is not robust to paraphrasing compared to other methods such as KGW, SemStamp, and PostMark.

\section{Experimental Settings} \label{appendix:eval-setting}
In all combinations of the experiments, following \citet{kgw}, sampling from the LLM was performed with a temperature of 0.7 and a repetition penalty of 1.05, while the minimum and the maximum number of generated tokens were set to 195 and 205, respectively. The maximum number of rejection sampling iterations was set to 100, again to align with the code provided by \citet{hou-etal-2024-semstamp, hou-etal-2024-k}. However, this setting reflects a trade-off between detection performance and generation speed. Based on our experiments, setting it to 25 achieves strong performance, with higher values offering only marginal improvements (see Appendix~\ref{appendix:max_trials}). For token-level watermarking baselines, in cases where results were not directly extracted from \citet{hou-etal-2024-semstamp, hou-etal-2024-k}, we employed the open-source MarkLLM watermarking framework \cite{pan-etal-2024-markllm}, with their recommended configurations ($\gamma=0.5, \delta=2$, \texttt{prefix\_length=1}, etc.) to run the experiments.

The majority of the experiments, including text generation and detection tasks, were conducted on a workstation equipped with an Intel Core i9 processor, 64GB of RAM, and an Nvidia RTX 3090 GPU with 24GB of VRAM. Some of the experiments involving bigram paraphrasing were performed on compute nodes with an Nvidia V100 GPU with 32GB of VRAM.

\section{Averaged Performance Across Datasets} \label{appendix:average-performance}

\begin{figure*}[!t]
    \centering
    \includegraphics[width=0.95\linewidth]{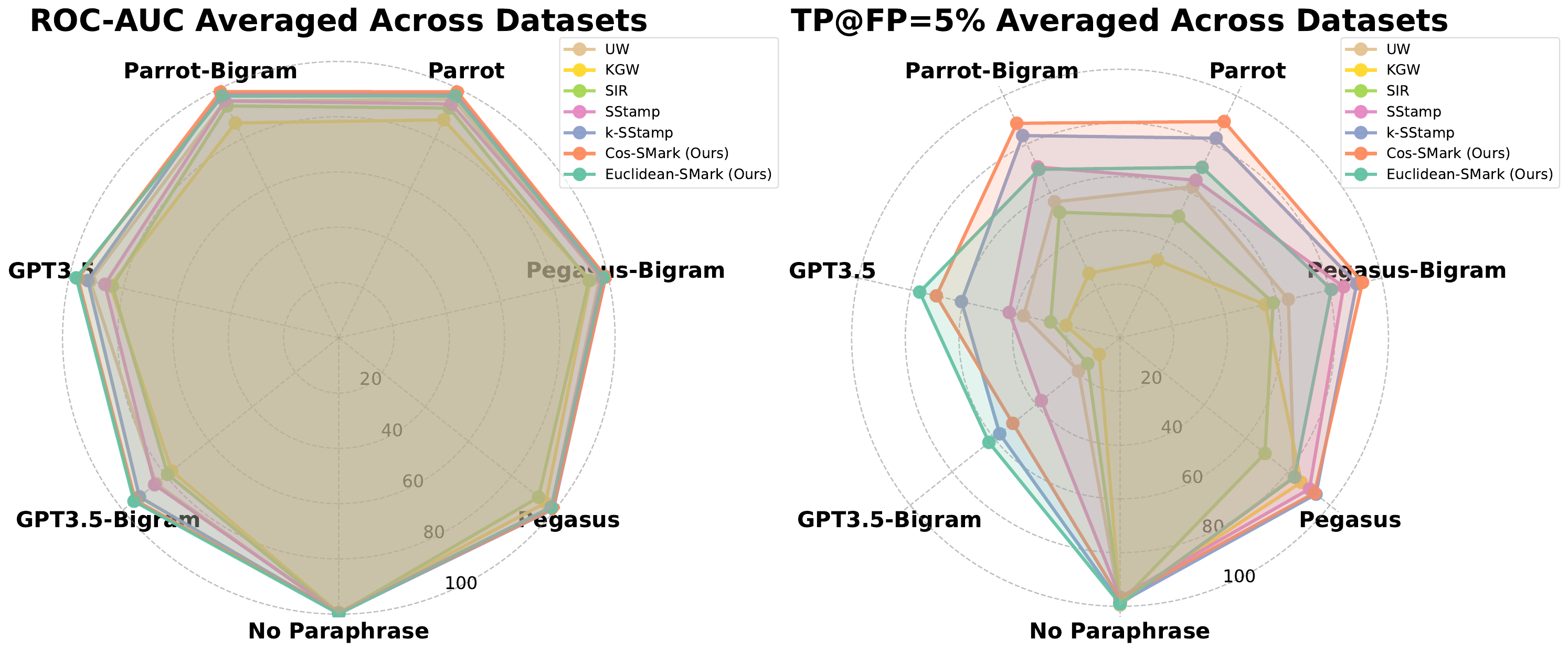}
    \caption{
    Detection performance averaged across three datasets.
    \textbf{Left:} ROC-AUC ↑ across different paraphraser variants. 
    \textbf{Right:} TP@FP=5\% ↑ under the same settings. 
    These radar plots provide a holistic comparison of all watermarking algorithms across paraphrasing conditions, averaged across datasets. 
    \textit{SimMark} demonstrates consistently strong robustness, closely matching or outperforming state-of-the-art baselines, particularly under heavier transformations.
    }
    \label{fig:radar2}
\end{figure*}

\begin{table*}[!t]
\centering
\resizebox{\textwidth}{!}{
\begin{tabular}{@{}clc|ccccccc@{}}
\toprule
Dataset & Algorithm & No Paraphrase & Pegasus & Pegasus-Bigram & Parrot & Parrot-Bigram & GPT3.5 & GPT3.5-bigram & Avg. Paraphrased \\
\midrule
\multirow{7}{*}{\rotatebox{90}{Avg. Over Datasets}}
 & UW & \textbf{99.9} / \textbf{99.5} / \textbf{99.9} & 98.4 / 82.9 / 95.0 & 96.9 / 64.3 / 87.3 & 95.8 / 62.5 / 83.6 & 95.1 / 56.2 / 80.9 & 92.3 / 36.9 / 66.2 & 84.7 / 19.8 / 43.8 & 93.8 / 53.8 / 76.1 \\
 & KGW & 99.5 / 98.3 / 98.7 & 95.8 / 86.3 / 91.2 & 93.4 / 55.5 / 75.3 & 87.5 / 32.1 / 58.2 & 86.3 / 26.7 / 53.6 & 84.8 / 20.6 / 49.9 & 77.3 / 9.9 / 31.6 & 87.5 / 38.5 / 60.0 \\
 & SIR & \underline{99.8} / 98.8 / \textbf{99.9} & 92.5 / 69.1 / 82.9 & 92.9 / 58.5 / 78.4 & 92.3 / 50.2 / 70.6 & 93.1 / 51.9 / 72.7 & 84.1 / 26.5 / 42.6 & 79.3 / 15.4 / 37.0 & 89.0 / 45.3 / 64.0 \\
 & SemStamp & 99.5 / 96.6 / 98.0 & 98.4 / 90.3 / 94.8 & 97.7 / 85.4 / 91.7 & 93.9 / 65.1 / 79.2 & 95.1 / 70.6 / 82.9 & 86.9 / 42.3 / 57.8 & 85.3 / 37.5 / 52.8 & 92.9 / 65.2 / 76.5 \\
 & $k$-SemStamp & \underline{99.8} / 98.6 / 99.1 & \textbf{99.4} / \textbf{93.4} / \underline{96.9} & \underline{99.0} / \underline{90.5} / \underline{95.6} & \underline{98.1} / \underline{82.5} / \underline{91.7} & \underline{98.2} / \underline{83.6} / \underline{91.1} & 93.2 / 60.6 / 77.4 & 92.3 / \underline{57.4} / 73.8 & 96.7 / \underline{77.8} / 87.7 \\
 & \textbf{Cosine-\textit{SimMark} (ours)} & 99.5 / 97.3 / 98.6 & \underline{99.2} / \underline{92.7} / \textbf{97.7} & \textbf{99.1} / \textbf{92.7} / \textbf{97.5} & \textbf{98.8} / \textbf{89.4} / \textbf{97.1} & \textbf{98.9} / \textbf{88.7} / \textbf{97.1} & \underline{96.9} / \underline{70.2} / \underline{89.4} & \underline{94.3} / 51.1 / \underline{78.8} & \textbf{97.8} / \textbf{80.8} / \textbf{92.9} \\
 & \textbf{Euclidean-\textit{SimMark} (ours)} & \textbf{99.9} / \underline{99.1} / \underline{99.5} & 98.3 / 83.2 / 93.9 & 98.2 / 80.8 / 92.3 & 97.1 / 70.5 / 87.7 & 97.1 / 69.6 / 87.8 & \textbf{97.5} / \textbf{76.6} / \textbf{90.1} & \textbf{94.8} / \textbf{62.5} / \textbf{79.5} & \underline{97.2} / 73.9 / \underline{88.6} \\
\bottomrule
\end{tabular}
}
\caption{Detailed average detection performance across the RealNews, BookSum, and Reddit-TIFU datasets under different paraphrasers. Each cell reports ROC-AUC ↑ / TP@1\%FP ↑ / TP@5\%FP ↑. Higher values indicate better performance across all metrics. \textbf{Bold} and \underline{underlined} numbers denote the highest and second-highest values, respectively. \textit{SimMark} consistently ranks among the top-performing methods in robustness across various paraphrasers averaged across datasets.}
  \label{tab:avg_performance_comparison}
\end{table*}

For a comprehensive comparison across all paraphrasing settings (as shown in Table \ref{tab:performance}), including ROC-AUC and TP@FP thresholds (1\% and 5\%), we report the averaged detection performance metrics across datasets in Table~\ref{tab:avg_performance_comparison}. To better visualize relative performance trends across paraphrasers, we also present radar plots in Figure~\ref{fig:radar2}, showing aggregated TP@FP=5\% and ROC-AUC scores (see Figure \ref{fig:radar1} in the main text for aggregated TP@FP=1\% scores). These visualizations complement the table by providing an intuitive view of robustness across settings. Together, these summaries highlight that \textit{SimMark} consistently outperforms or matches the baseline methods across diverse conditions.

\section{Additional Experimental Results}
\label{appendix:gemma3}

To demonstrate the model-agnostic nature of \textit{SimMark}, we applied our algorithm to the Gemma3-4B model\footnote{We employed \href{https://huggingface.co/google/gemma-3-4b-pt}{google/gemma-3-4b} (4B) model.} \cite{gemma3}. We evaluated both Cosine-\textit{SimMark} and Euclidean-\textit{SimMark} under different paraphrasing models across the same three datasets as before: RealNews subset of C4, BookSum, and Reddit-TIFU.
The effectiveness and robustness of watermarking techniques can depend heavily on the characteristics of the underlying LLM and the nature of the generated text. To maintain consistency and reliability across experiments, we made the following modifications:

\begin{itemize}[noitemsep, topsep=0pt]
    \item \textbf{Predefined Interval Adjustment:} The sentences' embedding similarity distribution under Gemma3-4B differed from those in OPT-1.3B, requiring new intervals. We set the predefined interval to [0.86, 0.90] for cosine similarity (without PCA), and [0.11, 0.16] for Euclidean distance (with PCA).
    \item \textbf{Threshold Transferability:} In contrast to our earlier experiments—where the detection threshold $\beta$ was determined per dataset—we fixed $\beta$ across all datasets in these experiments. Specifically, we determined the threshold using only non-watermarked data from the BookSum dataset, and then applied it across all three datasets without modification. This approach simulates a more realistic setting where the detector is calibrated on a single corpus but expected to generalize to others. The results demonstrate that our method maintains high detection performance even under this general configuration.
    \item \textbf{Longer Generations:} Since Gemma3-4B tends to generate longer sentences compared to OPT-1.3B model that we employed earlier, we increased the number of generated tokens from 200 to 300 to ensure a sufficient number of sentences for reliable hypothesis testing.
\end{itemize}
Table \ref{tab:gemma3} reports the detection performance in terms of ROC-AUC ↑ / TP@1\%FP ↑ / TP@5\%FP ↑, for each setting (↑: higher is better). Across all datasets and paraphrasing scenarios, \textit{SimMark} remains highly effective, with both cosine similarity and Euclidean distance variants maintaining strong ROC-AUC and TP rates. These results affirm that \textit{SimMark} maintains its performance across different LLM families (i.e., OPT and Gemma3) and datasets/domains, further validating the general applicability of our proposed algorithm.
\renewcommand*{\thefootnote}{\fnsymbol{footnote}}
\begin{table*}[t]
\centering
\resizebox{\textwidth}{!}{
\begin{tabular}{@{}clc|ccccccc@{}}
\toprule
Dataset & Method & No Paraphrase & Pegasus & Pegasus-Bigram & Parrot & Parrot-Bigram  & Avg. Paraphrased \\ 
\midrule
\multirow{2}{*}{\rotatebox{90}{RN}} 
& Cosine-\textit{SimMark} & 99.5 / 96.2 / 97.9 & 93.5 / 57.9 / 75.6 & 93.0 / 54.9 / 73.9 & 92.3 / 50.1 / 71.7 & 92.1 / 49.2 / 72.8 & 92.7 / 53.0 / 73.5\\
& Euclidean-\textit{SimMark}\footnotemark[7] & 99.5 / 97.4 / 98.2 & 93.2 / 65.3 / 79.5 & 91.8 / 61.8 / 78.5 & 91.9 / 58.0 / 73.9 & 91.5 / 58.7 / 72.8 & 92.1 / 61.0 / 76.2 \\
\midrule
\multirow{2}{*}{\rotatebox{90}{BS}} 
& Cosine-\textit{SimMark} & 99.9 / 99.6 / 99.8 & 97.7 / 75.8 / 90.5 & 97.4 / 73.3 / 90.0 & 97.0 / 67.8 / 87.6 & 96.9 / 67.7 / 86.7 & 97.2 / 71.1 / 88.7 \\
& Euclidean-\textit{SimMark}\footnotemark[7] & 99.9 / 99.7 / 99.9 & 97.6 / 82.0 / 91.4 & 97.4 / 77.9 / 89.9 & 97.4 / 78.0 / 91.2 & 91.5 / 58.7 / 72.8 & 96.0 / 74.2 / 86.3\\
\midrule
\multirow{2}{*}{\rotatebox{90}{TIFU}} 
& Cosine-\textit{SimMark} & 99.8 / 99.1 / 99.5 & 97.4 / 78.8 / 91.5 & 97.0 / 76.6 / 89.3 & 89.4 / 32.8 / 61.7 & 90.7 / 36.6 / 63.8 & 93.6 / 56.2 / 76.6\\
& Euclidean-\textit{SimMark}\footnotemark[7] & 99.6 / 98.8 / 99.1 & 97.3 / 82.0 / 91.4 & 96.9 / 81.0 / 90.1 & 92.2 / 53.6 / 73.9 & 92.8 / 58.4 / 76.0 & 94.8 / 68.8 / 82.8 \\
\bottomrule
\end{tabular}
}
\caption{
Performance of \textit{SimMark} using Gemma3-4B model across paraphrasers and datasets (RealNews denoted as RN, BookSum denoted as BS, and Reddit-TIFU denoted as TIFU). Each cell reports AUC $\uparrow$ / TP@FP=1\% $\uparrow$ / TP@FP=5\% $\uparrow$ (↑: higher is better). The results demonstrate that \textit{SimMark} maintains strong performance across different datasets and paraphrasing conditions, highlighting its robustness and model-agnostic nature.
}
\label{tab:gemma3}
\end{table*}

\section{Effect of Sampling Budget on Detection Performance}
\label{appendix:max_trials}

To better understand the trade-off between generation speed and detection performance, we analyze the impact of the \texttt{max\_trials} hyperparameter, which defines the upper limit on the number of rejection sampling iterations during watermark injection. While we set this value to 100 in our main experiments (to align with prior works of \citet{hou-etal-2024-semstamp, hou-etal-2024-k}), it is important to examine whether such a large value is necessary.

\begin{figure}[ht]
    \centering
    \includegraphics[width=\linewidth]{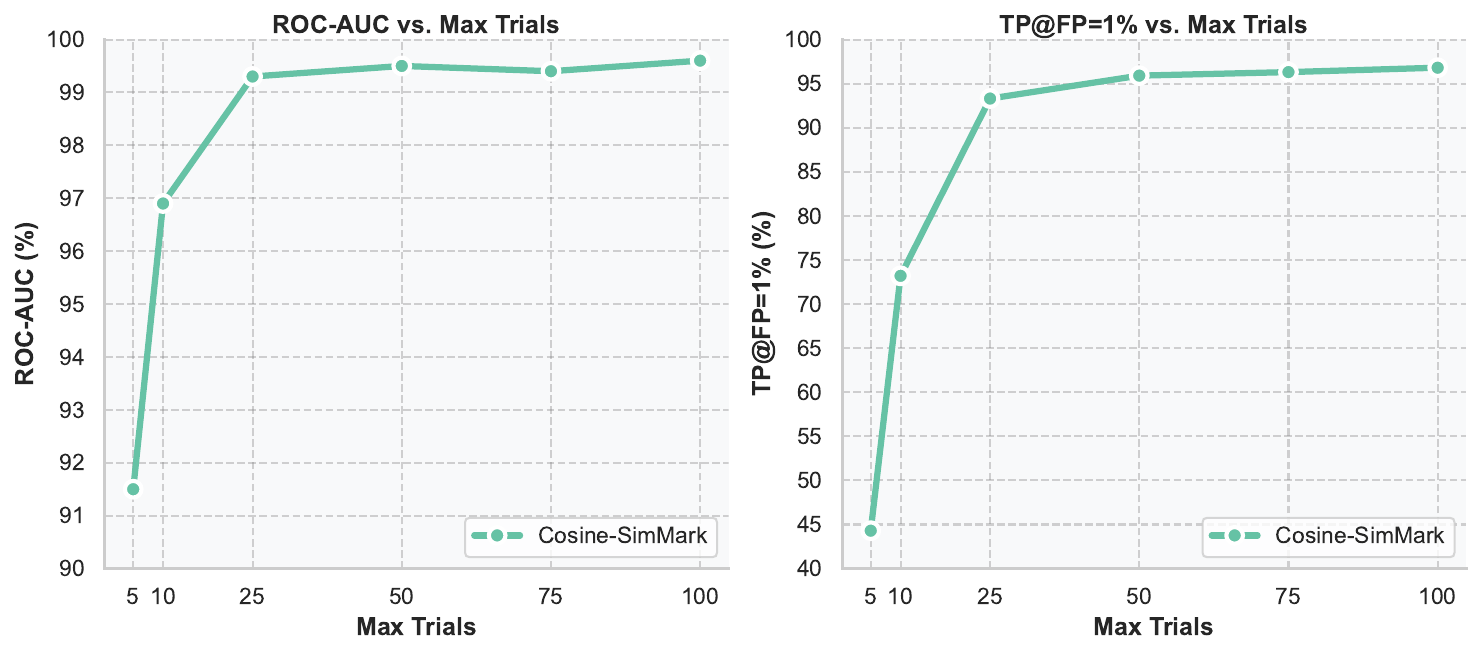} 
    \caption{
    Impact of the maximum number of rejection sampling trials on detection performance. 
    Increasing \texttt{max\_trials} improves both ROC-AUC ↑ and TP@1\%FP ↑ (↑: higher is better), but the improvement plateaus around 25. 
    Results are reported on the RealNews dataset using cosine-\textit{SimMark} and OPT-1.3B model.
    }
    \label{fig:max_trials}
\end{figure}

Figure~\ref{fig:max_trials} shows two evaluation metrics—ROC-AUC ↑ and TP@FP=1\% ↑ (↑: higher is better)—on the RealNews dataset for cosine-\textit{SimMark} and OPT-1.3B model under different values of \texttt{max\_trials}. As shown in the plots, performance improves significantly when increasing \texttt{max\_trials} from 5 to 25, but plateaus thereafter. In particular, both ROC-AUC and TP@FP=1\% show diminishing returns beyond 25 trials, indicating that additional sampling brings little performance gain.

These results suggest that setting \texttt{max\_trials} to 25 achieves a good balance between robustness and efficiency, and using larger values (e.g., 100) is not strictly necessary in practice. These findings, together with the average sampling statistics reported in the main paper (e.g., 7.1 samples per sentence), highlight \textit{SimMark}’s ability to balance robustness with generation speed compared to SemStamp (20.9 samples per sentence) and $k$-SemStamp (13.3 samples per sentence).

\section{Empirical Analysis of Generation Efficiency} \label{appendix:generation-efficiency}

We experimentally evaluate the generation-time latency introduced by \textit{SimMark}. Using 100 samples from the BookSum dataset, we measure the per-sentence generation time under various configurations of \texttt{max\_trials}, using OPT-1.3B model. On our hardware (see Appendix \ref{appendix:eval-setting}), with \texttt{max\_trials} = 100 (i.e., maximum number of rejection sampling), cosine-\textit{SimMark} yields an average latency of 1.33 seconds per sentence—corresponding to a 7.1× slowdown compared to the unwatermarked baseline average (0.188 sec/sentence). This matches our reported average sampling rate of 7.1 candidates per sentence (see Section~\ref{experiments}).

\footnotetext[7]{PCA is applied.}
\renewcommand*{\thefootnote}{\arabic{footnote}}

For reference, token-level methods such as UW and SIR yield 0.182 and 0.377 seconds per sentence, respectively (i.e., around 1× and 2× slowdown). This highlights the key trade-off between efficiency and robustness: while token-level methods are faster, \textit{SimMark} offers less impact on text quality and stronger resistance to paraphrasing.

\begin{table}[!ht]
\centering
\resizebox{0.48\textwidth}{!}{
\begin{tabular}{lcc}
\toprule
Method & Latency & Slowdown \\
\midrule
Baseline (no watermark) & 0.188 sec/sent & 1.0× \\
\textit{SimMark} (\texttt{max\_trials}=25) & 0.977 sec/sent &  5.2× \\
\textit{SimMark} (\texttt{max\_trials}=50) & 1.114 sec/sent & 5.9× \\
\textit{SimMark} (\texttt{max\_trials}=100) & 1.330 sec/sent & 7.1× \\
UW (token-level) & 0.182 sec/sent & $\sim$1.0× \\
SIR (token-level) & 0.377 sec/sent & 2.0× \\
\bottomrule
\end{tabular}
}
\caption{Latency comparison of different watermarking methods (in seconds per sentence) and their slowdowns relative to unwatermarked generation. \textit{SimMark}'s higher overhead stems from rejection sampling but remains practical—especially at lower \texttt{max\_trials}—and can be further reduced with efficient decoding backends (e.g., vLLM).}
\label{tab:latency}
\end{table}

In table \ref{tab:latency}, we also evaluate \textit{SimMark} under smaller \texttt{max\_trials} values for practical trade-offs, as performance remains strong even with a smaller sampling budget (see Appendix \ref{appendix:max_trials}).

While \textit{SimMark} introduces overhead due to rejection sampling, newer decoding backends (e.g., vLLM \cite{kwon2023efficient}) can reduce this cost substantially via prompt-prefix reuse and KV cache optimization. However, our current benchmarks were conducted using the Hugging Face Transformers library \cite{wolf-etal-2020-transformers} without these optimizations.

 Although we were unable to directly benchmark SemStamp and $k$-SemStamp due to technical issues, their substantially higher sampling requirements suggest that they would incur 2–3× greater overhead than \textit{SimMark}. It is worth noting that a generation latency of about 1 second per sentence remains below the average human reading speed (200–250 wpm, or roughly 3–6 seconds per sentence). In summary, we believe \textit{SimMark} offers a reasonable trade-off between detection robustness and runtime overhead, maintaining practicality for many real-world applications even in long-form scenarios.

\section{Theoretical Analysis of Sampling Efficiency}  \label{appendix:sampling-efficiency}

The average number of samples required to generate a valid sentence is influenced by the chosen interval \([a, b]\). To provide further insights into this relationship, we estimated the area under the curve (AUC) of the embedding similarity distribution for an unwatermarked LLM (OPT-1.3B). For instance, for the interval \([0.68, 0.76]\) (for cosine-\textit{SimMark}), the estimated AUC is approximately \( 0.194 \). 

Using the mean of the geometric distribution, which is given by \( \frac{1}{p} \), where \( p \) is the probability of success (in this case, the probability of falling within the interval), this translates to an expected average of $\frac{1}{0.194} \approx 5.1$ samples per valid sentence. This estimate is on par with the experimental results reported in the main paper. The AUC estimates were computed using the binning technique, as described in Appendix \ref{appendix:estimate-AUC}.

This analysis underscores the importance of carefully selecting the interval \([a, b]\), as narrower intervals may increase the number of required samples, leading to reduced sampling efficiency but better performance, while broader intervals may compromise the effectiveness of the watermark. By understanding the interplay between the interval choice and sampling efficiency, we can better optimize \textit{SimMark}'s performance.

\section{Examples of Watermarked Text} \label{appendix:examples}
\begin{figure*}[!h]
    \centering
    \includegraphics[page=3, width=\linewidth]{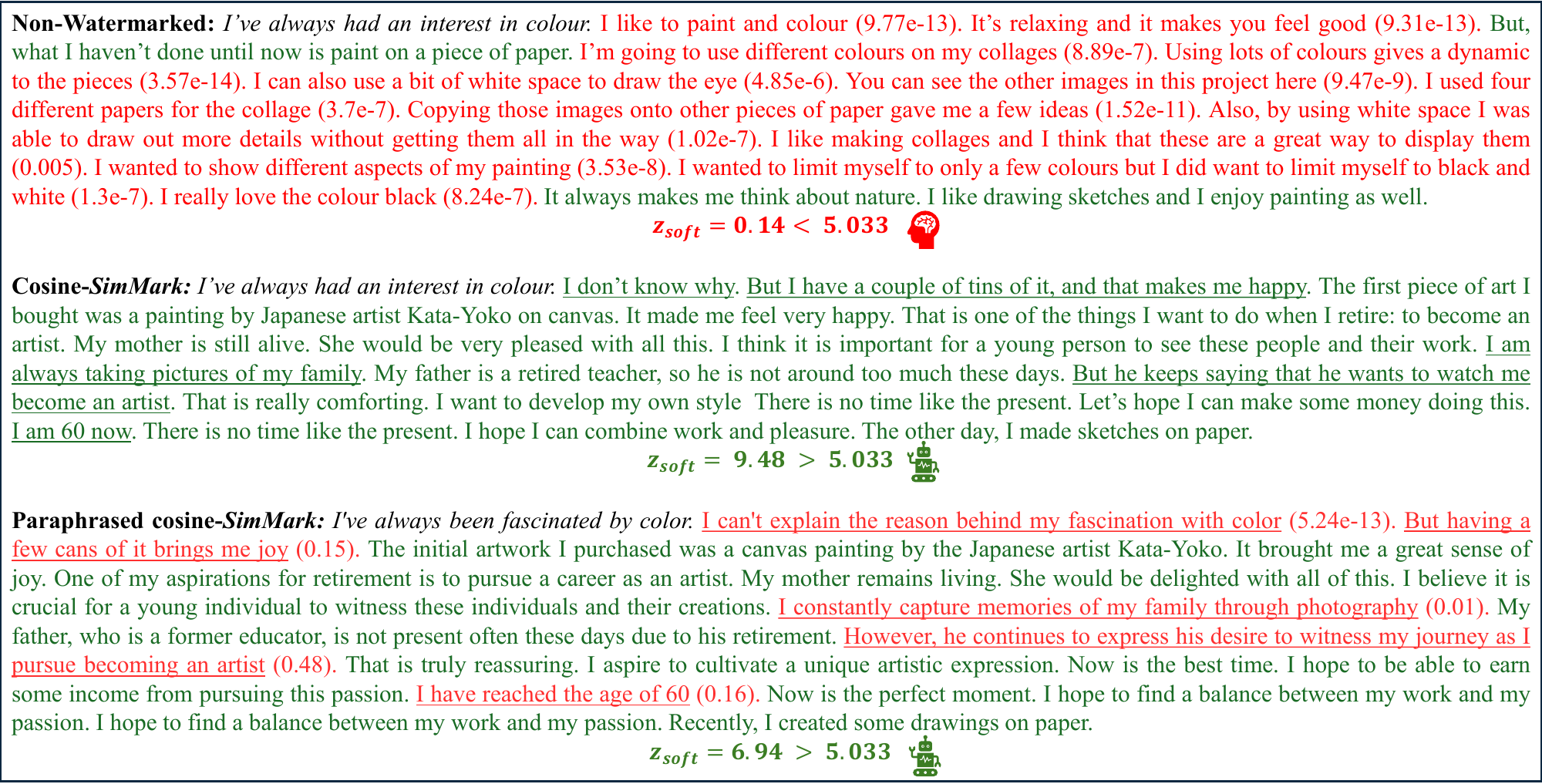}
    \caption{Example of text generated with and without cosine-\textit{SimMark} using RealNews dataset and OPT-1.3B model. The first sentence (in black) is the prompt for the model, the \textcolor{green!50!black}{green} sentences are \textcolor{green!50!black}{valid}, and \textcolor{Red}{red} sentences are \textcolor{Red}{invalid/partially valid}. Numbers in parentheses represent the \textit{soft count} for partially valid sentences. The top panel shows non-watermarked text, which fails to produce a significant detection signal (\( z_{\text{soft}} = 0.14 \; \textcolor{Red}{<} \; 5.033 \), false negative). The middle panel demonstrates text generated using \textit{SimMark} with cosine similarity-based watermarking, producing a strong detection signal (\( z_{\text{soft}} = 9.48 \; \textcolor{green!50!black}{>} \; 5.033 \)). The bottom panel shows paraphrased watermarked text using GPT-3.5-Turbo, where the embedded watermark remains detectable despite paraphrasing (\( z_{\text{soft}} = 6.94 \; \textcolor{green!50!black}{>} \; 5.033 \)).}
    \label{fig:ex1}
\end{figure*}

\begin{figure*}[h]
    \centering
    \includegraphics[page=4, width=\linewidth]{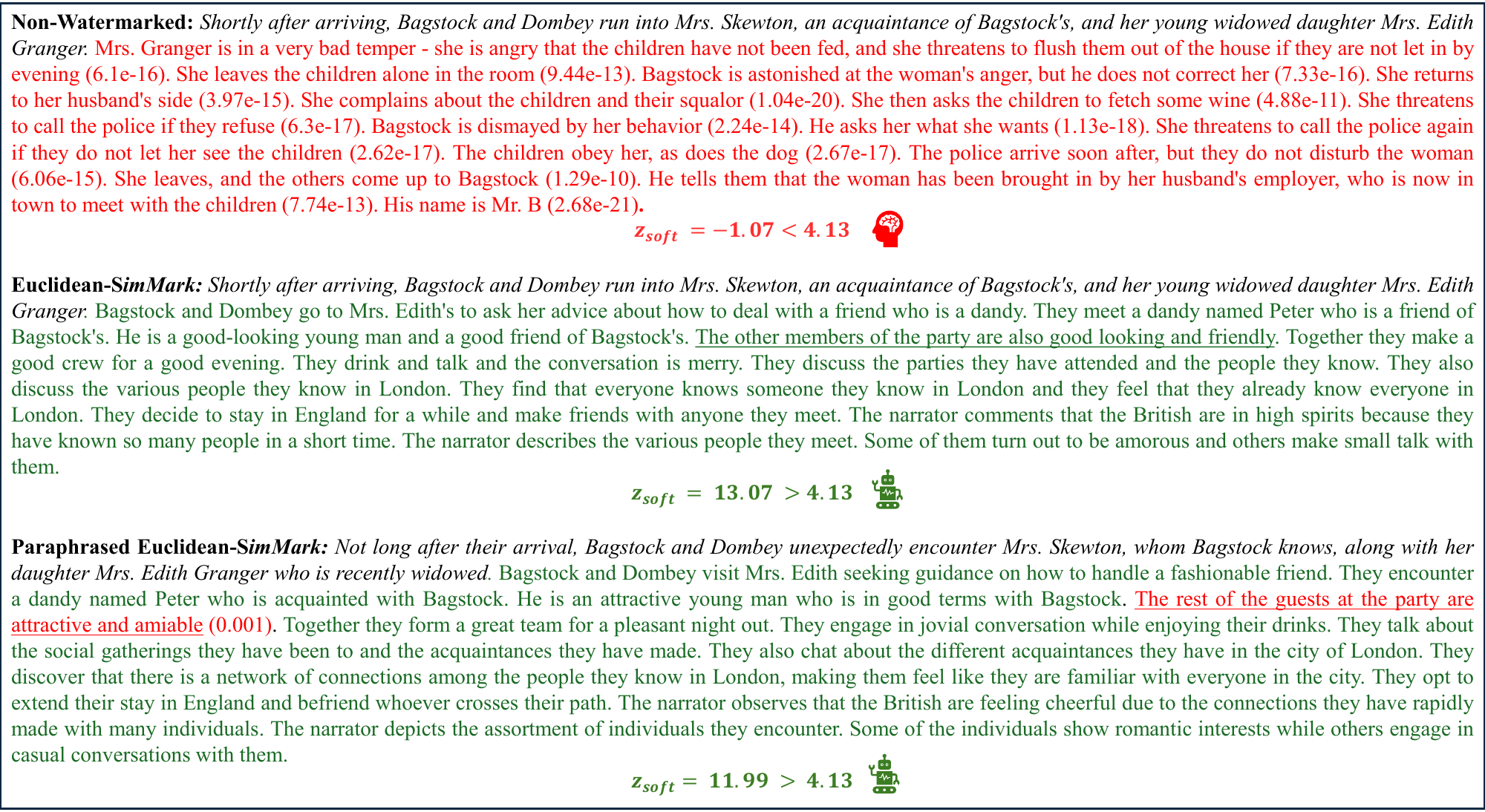}
    \caption{Example of text generated with and without Euclidean-\textit{SimMark} using BookSum dataset and OPT-1.3B model. The first sentence (in black) is the prompt for the model, the \textcolor{green!50!black}{green} sentences are \textcolor{green!50!black}{valid}, and \textcolor{Red}{red} sentences are \textcolor{Red}{invalid/partially valid}. Numbers in parentheses represent the \textit{soft count} for partially valid sentences. The top panel shows the non-watermarked text, which fails to produce a significant detection signal (\( z_{\text{soft}} = -1.07 \; \textcolor{Red}{<} \; 4.13 \), false negative). The middle panel demonstrates text generated using \textit{SimMark} with Euclidean distance-based watermarking, producing a strong detection signal (\( z_{\text{soft}} = 13.07 \; \textcolor{green!50!black}{>} \; 4.13 \)). The bottom panel shows paraphrased watermarked text using GPT-3.5-Turbo, where the embedded watermark remains detectable despite paraphrasing (\( z_{\text{soft}} = 11.99 \; \textcolor{green!50!black}{>} 
\; 4.13 \)).}
    \label{fig:ex2}
\end{figure*}

Figures~\ref{fig:ex1} and~\ref{fig:ex2} provide examples of text generated with and without the \textit{SimMark} watermark using OPT-1.3B. These examples illustrate the imperceptibility of the watermark to human readers while enabling robust detection through our proposed algorithm. They also highlight \textit{SimMark}'s robustness to paraphrasing while maintaining quality comparable to non-watermarked text.  

\section{Ablation Study on Soft Count Smoothness Factor $K$} \label{appendix:K}
\begin{table*}[h]
\centering
 \resizebox{0.98\textwidth}{!}{
\begin{tabular}{llcc|cc} 
\toprule 
Count Method & K & Cosine-\textit{SimMark} & Paraphrased cosine-\textit{SimMark} & Euclidean-\textit{SimMark} & Paraphrased Euclidean-\textit{SimMark} \\
\midrule 
\multirow{4}{*}{Soft Count} & 50 & 99.0 / 89.2 / 97.2 & \underline{98.6} / 78.5 / 96.5 & \underline{99.4} / 91.2 / 98.1 & 96.9 / 49.3 / 87.9 \\
& 150 & \underline{99.6} / \underline{95.7} / \underline{98.8} & \textbf{99.2} / 88.7 / \textbf{98.2} & \textbf{99.8} / \underline{97.6} / \underline{99.3} & \textbf{97.3} / 67.8 / \textbf{90.4} \\
& 250 & \textbf{99.7} / \textbf{96.9} / \underline{98.8} & \textbf{99.2} / \underline{90.3} / \textbf{98.2} & \textbf{99.8} / \textbf{98.5} / 99.2 & \underline{97.2} / \textbf{72.3} / \underline{88.9} \\
& 350 & \textbf{99.7} / \textbf{96.9} / \textbf{98.9} & \textbf{99.2} / \textbf{90.4} / \underline{98.1} & \textbf{99.8} / \textbf{98.5} / \textbf{99.4} & \underline{97.2} / \underline{71.1} / \underline{88.9} \\ 
\midrule
\multirow{1}{*}{Regular Count}
& $\infty$ & 99.7 / 97.2 / 99.1 & 99.1 / 88.7 / 97.6 & 99.8 / 98.5 / 99.7 & 97.0 / 70.0 / 88.2 \\ 
\bottomrule 
\end{tabular}
}
\caption{Ablation study on the smoothness factor \( K \) in soft counting (Eq.~\eqref{eq:count_formula}) using the RealNews dataset, with Pegasus as the paraphraser. Metrics reported include ROC-AUC $\uparrow$, TP@FP=1\% $\uparrow$, and TP@FP=5\% $\uparrow$, from left to right. The last row (\( K = \infty \)) corresponds to regular counting with a step function in the interval \([a, b]\). A smoothness factor of \( K=250 \) provides a good balance between performance before and after paraphrase attacks for both cosine-\textit{SimMark} and Euclidean-\textit{SimMark}. Notably, while soft counting slightly reduces performance in the absence of paraphrasing, it demonstrates enhanced robustness against paraphrasing, yielding an increase across all metrics for Pegasus paraphraser and potentially larger gains against more advanced paraphrasers.}
\label{tab:K}
\end{table*}
In this section, we analyze the impact of the smoothness factor \( K \) on the performance of \textit{SimMark}. Recall that \( K \) controls the degree of smoothness in the soft counting mechanism as defined in Eq.~\eqref{eq:count_formula}. A larger \( K \) makes the soft counting function behave more like a step function, while smaller values provide smoother transitions between valid and invalid sentences.

Table~\ref{tab:K} presents the results of this ablation study, conducted on the RealNews dataset with Pegasus as the paraphraser. Metrics include ROC-AUC, TP@FP=1\%, and TP@FP=5\%. Higher values indicate better performance across all metrics. The results demonstrate the following trends:
\begin{itemize}[noitemsep, topsep=0pt]
    \item A smoothness factor of \( K=250 \) provides a good trade-off, achieving strong performance both before and after paraphrasing attacks for both cosine-\textit{SimMark} and Euclidean-\textit{SimMark}.
    \item For \( K=\infty \), corresponding to regular counting with a step function, the performance is slightly higher in the absence of paraphrasing but significantly degrades under paraphrasing attacks, highlighting the benefits of soft counting in adversarial scenarios.
\end{itemize}

These findings confirm that soft counting loses a small amount of performance when no paraphrasing is applied, but it gains substantial robustness under paraphrasing. For example, TP@FP=1\% improves by 1.6–2.3\% for Pegasus-paraphrased text when \( K=250 \), and the improvement is likely to be even more significant for stronger paraphrasers.

\section{Ablation Study on Impact of PCA} \label{appendix:ablation_pca}
Table \ref{tab:pca} presents the results of an ablation study investigating the impact of applying PCA to reduce the dimensionality of sentence embeddings across RealNews, BookSum, and Reddit-TIFU datasets. Metrics include ROC-AUC, and TP at fixed FP rates (FP=1\% and FP=5\%). Higher values indicate better performance across all metrics, with PCA applied to embeddings to explore its effect on detection accuracy and robustness. 

The results reveal that the effect of PCA depends on the choice of similarity measure. For \textit{Euclidean distance-based SimMark}, applying PCA generally improves robustness against paraphrasing attacks across most datasets, except for the BookSum dataset. This improvement likely arises because reducing dimensionality helps mitigate noise in the embeddings, especially after the paraphrasing attack.
On the other hand, for \textit{cosine similarity-based} \textit{SimMark}, applying PCA reduces performance across all datasets. This reduction may be due to PCA altering the embeddings in a way that disrupts the angular relationships critical for cosine similarity calculations. These findings highlight the importance of adapting PCA usage based on the similarity measure employed to achieve optimal watermarking performance.

\begin{table}[t]
\centering
\resizebox{0.48\textwidth}{!}{
\begin{tabular}{@{}clcc@{}}
\toprule
Dataset & Configuration   & No paraphrase & Pegasus \\ 
\midrule
\multirow{4.5}{*}{\rotatebox{90}{RealNews}} 
& Cosine-\textit{SimMark} (No PCA)                 & \textbf{99.7} / 96.9 / 98.8                      & \textbf{99.2} / \textbf{90.3} / \textbf{98.2}                                             \\
& Cosine-\textit{SimMark} (PCA)                     & 99.6 / 96.9 / \textbf{99.1}                      & 92.1 / 33.8 / 71.2                                             \\
\cmidrule{2-4}
& Euclidean-\textit{SimMark} (No PCA)               & 99.4 / 92.6 / 98.4                      & 90.5 / 19.7 / 58.0                                              \\
& Euclidean-\textit{SimMark} (PCA)                 & \textbf{99.8} / \textbf{98.5} / \textbf{99.2}                      & \textbf{97.2} / \textbf{72.3} / \textbf{88.9}                                             \\ 
\midrule 
\multirow{4.5}{*}{\rotatebox{90}{BookSum}} 
& Cosine-\textit{SimMark} (No PCA)            & 99.8 / 98.8 / 99.5                      & \textbf{99.5} / \textbf{93.3} / \textbf{98.5}                                              \\
& Cosine-\textit{SimMark} (PCA)               & \textbf{100} / \textbf{99.9} / \textbf{99.9}                     & 98.7 / 87.3 / 95.1                                              \\ 
\cmidrule{2-4}
& Euclidean-\textit{SimMark} (No PCA)         & \textbf{100} / \textbf{100} / \textbf{100}                      & \textbf{98.8} / \textbf{82.6} / \textbf{94.9}                                             \\
& Euclidean-\textit{SimMark} (PCA)            & 99.9 / 99.3 / 99.5                     & 97.4 / 69.8 / 88.6                                              \\ 
\midrule 
\multirow{4.5}{*}{\rotatebox{90}{\small Reddit-TIFU}} 
& Cosine-\textit{SimMark} (No PCA)              & 99.1 / 96.3 / 97.6          & \textbf{98.9} / \textbf{94.5} / \textbf{96.4} \\
& Cosine-\textit{SimMark} (PCA)              & \textbf{99.7} / \textbf{98.8 }/ \textbf{99.3}                      & 96.6 / 78.9 / 89.3  \\
\cmidrule{2-4}
& Euclidean-\textit{SimMark} (No PCA)              & 99.6 / 98.1 / 99.1                      & 96.7 / 72.6 / 90.0 \\
& Euclidean-\textit{SimMark} (PCA)              & \textbf{99.8} / \textbf{98.7} / \textbf{99.2}                      & \textbf{99.0} / \textbf{94.7} / \textbf{97.6} \\
\bottomrule
\end{tabular}
 }
\caption{Ablation study on the impact of applying PCA to embeddings across three datasets. Metrics reported include ROC-AUC $\uparrow$, TP@FP=1\% $\uparrow$, and TP@FP=5\% $\uparrow$, respectively, from left to right. Higher values indicate better performance across all metrics. \textbf{Bold} and \underline{underlined} numbers denote the highest and second-highest values, respectively. For cosine-\textit{SimMark}, not applying PCA yields better results, while for Euclidean-\textit{SimMark}, applying PCA improves performance except on the BookSum dataset.}
\label{tab:pca}
\end{table}

\section{Finding an Optimal Interval} \label{appendix:optimal_interval}

\begin{figure}[t]
    \centering
\includegraphics[width=\linewidth]{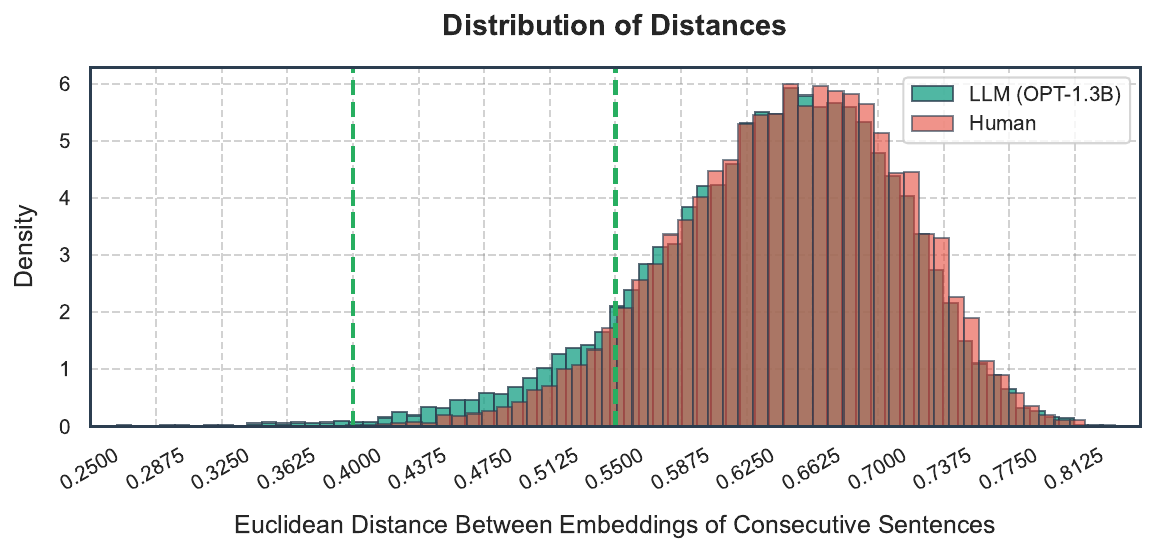}
    \caption{Distribution of Euclidean distances between embeddings of consecutive sentences for \textcolor{Salmon}{human-written} and \textcolor{SeaGreen}{LLM-generated} text on BookSum dataset, generated using OPT-1.3B. The figure demonstrates that the interval $[0.4, 0.55]$ is a reasonable choice for Euclidean-\textit{SimMark} in this case, though it is not necessarily the only viable option.}
    \label{fig:dist_of_distances}
\end{figure}

Figure~\ref{fig:dist_of_distances} shows the distribution of distances between embeddings of consecutive sentences for both human and LLM-generated text, calculated on a sample of size 1000 from the BookSum dataset (no PCA applied to the embeddings in this case). A small but noticeable distribution shift between the two can be observed. Based on this, the interval \([0.4, 0.55]\) appears to be a reasonable choice for \textit{SimMark} in this case. It is important to note that changes to the embedding representations, such as applying PCA or using a different embedding model, will lead to altered distance distributions. Consequently, the interval must be adjusted accordingly to maintain optimal performance. For instance, if PCA is applied, the interval \([0.28, 0.36]\) is suitable. Similarly, if we plot the figure for when cosine similarity is used instead of Euclidean distance, intervals \([0.81, 0.94]\)  and \([0.68, 0.76]\) are good candidates for cases with and without PCA, respectively. This variability in the distance distributions may also strengthen the algorithm's resistance to reverse engineering.

Selecting the optimal interval \([a, b]\) is a critical step in achieving a robust and reliable watermarking with \textit{SimMark}. In general, selecting an optimal interval involves balancing low FP rates, high TP rates, and robustness against paraphrasing attacks. It is often beneficial to choose intervals toward the tails of the distribution rather than around the mean. Finally, further exploration of dynamic interval selection mechanisms could enhance \textit{SimMark}'s robustness.

\section{Computing Threshold \( \beta \) for \textit{soft}-$z$-test} \label{appendix:estimate-AUC}
Recall that a text is classified as LLM-generated when \textcolor{green!50!black}{$z_\mathrm{soft} > \beta$}, and as human-written otherwise. $z_\mathrm{soft}$ is the statistic used in the statistical test described in Eq. \eqref{eq:z_formula2}. To determine the threshold \( \beta \) that limits the FP rate to 1\% or 5\%, we first need to estimate \( p_0 \), the probability that the consecutive embeddings' similarity or distance falls within the predefined interval \([a, b]\). This value of \( p_0 \) is a key component in calculating the $z_\mathrm{soft}$, as it represents the proportion of valid sentences in human-written text under the given interval. \( p_0 \) serves as an indicator of how frequently \textit{valid} sentences are expected to occur in human-authored text.

To compute \( p_0 \), we analyze the distribution of similarities (or distances) using a histogram approach, such as the one depicted in Figure~\ref{fig:dist_of_distances}. Specifically, we employ a binning technique to approximate the area under the curve of distribution in the interval \([a, b]\). The process involves dividing the entire range of distances or similarities into a fixed number of bins—1000 bins in our implementation. Each bin represents a small segment of the range, and the histogram is used to calculate the proportion of samples that fall within the interval \([a, b]\). Mathematically, \( p_0 \) is estimated as:

{\small\[
p_0 = \frac{\text{Number of samples in bins corresponding to } [a, b]}{\text{Size of the dataset}},
\]}once \( p_0 \) is estimated, the detection threshold \( \beta \) is determined by iterating over a range of possible values, typically from -10 to 10, to find the one that results in the desired false positive rate. Specifically, the threshold is chosen such that the proportion of human-written texts misclassified as LLM-generated matches the target FP rate (e.g., 1\% or 5\%).

It is worth noting that the distribution of similarities or distances may vary depending on different factors such as the embedding model and similarity measure (e.g., cosine or Euclidean). As a result, \( p_0 \) and therefore \( \beta \) are determined programmatically during the detection to ensure reliable performance of the watermarking algorithm.

\section{Human Evaluation of \textit{SimMark}'s Imperceptibility} \label{appendix:human_eval}

\begin{figure*}[t]
    \centering
    \includegraphics[width=0.95\textwidth]{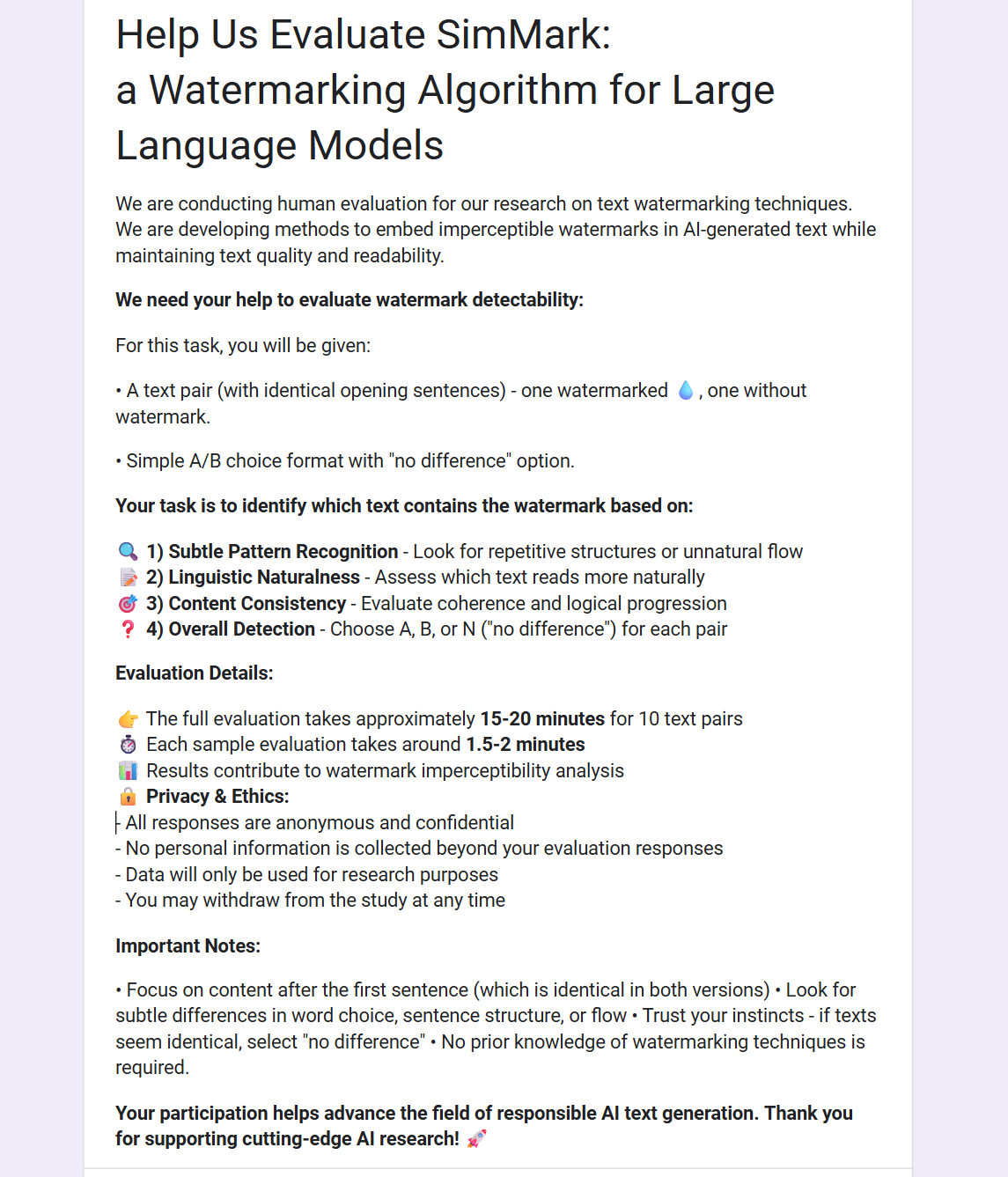}
    \caption{Screenshot of the instructions and consent form shown to participants before the human evaluation study. It describes the task, evaluation criteria, expected time commitment, and privacy assurances.}
    \label{fig:human_eval_form}
\end{figure*}

Human evaluation provides a more direct measure of imperceptibility for our watermarking algorithm. To assess this, we conducted a small-scale A/B test comparing outputs generated by \textit{SimMark} to unwatermarked ones, using 10 randomly selected BookSum samples. Each sample pair consisted of one watermarked and one unwatermarked text.

\begin{table}[!ht] 
\centering
\small
\resizebox{0.48\textwidth}{!}{
\begin{tabular}{lc}
\toprule
Option Chosen & Percentage \\
\midrule
\textit{SimMark} identified as watermarked & 36.7\% \\
No noticeable difference & 23.3\% \\
Unwatermarked identified as watermarked & 40.0\% \\
\bottomrule
\end{tabular}
}
\caption{Results of a small-scale human evaluation measuring the imperceptibility of \textit{SimMark}. Participants were asked to identify which of two outputs (watermarked vs. unwatermarked) appeared to be watermarked across 10 random samples. Responses were nearly evenly split, indicating that \textit{SimMark} watermarking is largely imperceptible to human readers.}
\label{tab:human_eval}
\end{table}

Three volunteers (all master's students in Canada, Iranian, aged between 18 and 25, with a Computer Science background) participated in the study. No sensitive data was collected beyond age range, academic level, and general field of study. Before starting the evaluation, participants were shown the consent and instruction form (Figure~\ref{fig:human_eval_form}), which described the task, the absence of any risks, and the use of their responses solely for academic purposes. Participants provided explicit consent by filling out the form and then proceeded with the evaluation task. 

After filling in the form, participants were asked to identify which option—\textit{A} or \textit{B}—was more likely to contain a watermark. We also included the ``\textit{N}” (no noticeable difference) option to avoid forcing a binary choice when participants genuinely could not distinguish between the texts, thereby making the evaluation more reliable given the scale of the evaluation.

\begin{table}[!t]
    \centering
    \small
    \begin{tabular}{p{7cm}}
    \hline
    \multicolumn{1}{c}{\textbf{Prompt for Regular Attack}} \\
    \hline
    \texttt{Previous context: \{context\} \textbackslash n Current sentence to paraphrase: \{sent\}} \\
    \hline
    \multicolumn{1}{c}{\textbf{Prompt for Bigram Attack}} \\
    \hline
    \texttt{Previous context: \{context\} \textbackslash n Paraphrase in \{num\_beams\} different ways and return a numbered list: \{sent\}} \\
    \hline
    \end{tabular}
    \caption{Prompts used to generate paraphrases with GPT-3.5-Turbo for regular and bigram attacks. These are the same prompts used by \citet{hou-etal-2024-semstamp} for consistent and comparable evaluation. Here, ``\texttt{sent}'' represents the target sentence to rephrase, ``\texttt{context}'' includes all preceding sentences, and ``\texttt{num\_beams}'' specifies the number of paraphrases generated for the bigram attack. A higher ``\texttt{num\_beams}'' value indicates a more aggressive attack. Following \citet{hou-etal-2024-semstamp}, we set ``\texttt{num\_beams}'' to 10 to have 10 rephrases of each sentence.}
    \label{tab:prompts}
\end{table}

We focused our human evaluation on asking participants to detect watermarking, rather than assess fluency or preference. This directly targets \textit{SimMark}’s design objective of imperceptibility: ensuring that even when readers are explicitly prompted to spot a watermark, they find it difficult to do so.

As shown in Table \ref{tab:human_eval}, on average, participants labeled the \textit{SimMark} output as watermarked in 36.7\% of cases, while in 40.0\% of cases they mistakenly identified the unwatermarked text as watermarked. In 23.3\% of comparisons, they reported no noticeable difference. These results are close to random guess, suggesting that \textit{SimMark}'s watermark is largely imperceptible to human readers while maintaining natural fluency. A larger-scale evaluation is deferred to future work.

\section{Prompts Used with GPT-3.5-Turbo for Paraphrasing} \label{appendix:prompt}
 Table \ref{tab:prompts} presents the prompts we used to obtain paraphrases using GPT-3.5-Turbo (accessed via OpenAI API\footnote{\href{https://platform.openai.com/docs/api-reference}{https://platform.openai.com/docs/api-reference}}) for both regular paraphrasing and more aggressive bigram paraphrasing attacks\footnote{Used ``\texttt{gpt-3.5-turbo-16k}'' model.}. By using the same prompts as \citet{hou-etal-2024-semstamp}, we ensured that our results were directly comparable to those extracted from their paper, maintaining consistency in evaluation methodology.
\end{document}